\documentclass[runningheads]{llncs}

\usepackage{eccv}

\usepackage{eccvabbrv}

\usepackage{graphicx}
\usepackage{booktabs}
\usepackage{arydshln}

\usepackage[accsupp]{axessibility}  

\usepackage{pifont}
\usepackage{makecell}
\usepackage{placeins}

\renewcommand{\paragraph}[1]{\vspace{1mm}\noindent\textbf{#1}}

\AtEndPreamble{
    \usepackage[capitalize]{cleveref}
    \crefname{section}{Sec.}{Secs.}
    \Crefname{section}{Section}{Sections}
    \Crefname{table}{Table}{Tables}
    \crefname{table}{Tab.}{Tabs.}
}

\usepackage[pagebackref,breaklinks,colorlinks,citecolor=eccvblue]{hyperref}

\usepackage{orcidlink}
\usepackage{color}
\usepackage{multirow}
\usepackage{colortbl} 
\usepackage{enumitem}

\definecolor{best}{HTML}{FFE7B1}
\definecolor{shade1}{HTML}{9FD77F}
\definecolor{shade1-1}{HTML}{D4FBD9}
\definecolor{shade2}{HTML}{F4B4B3}

\definecolor{shade2-1}{HTML}{FBDCD0}

\definecolor{shade3}{HTML}{CAEEFB}
\definecolor{shade3-1}{HTML}{E0F8FB}
\definecolor{shade3-1-1}{HTML}{CEF8FB}
\definecolor{shade3-2}{HTML}{CEF8FB}
\definecolor{shade3-2-1}{HTML}{E0F8FB}

\definecolor{shade4}{HTML}{ECD0EC}
\definecolor{shade4-1}{HTML}{EFE4F1}
\definecolor{shade5}{HTML}{C0C5EC}
\definecolor{shade5-1}{HTML}{E7E4F9}

\begin{document}

\title{RawGen: Learning Camera Raw \\ Image Generation
\vspace{-5mm}
}


\author{Dongyoung Kim$^1$\orcidlink{0009-0000-6414-2380} \and
Junyong Lee$^{1*}$\orcidlink{0000-0001-6472-0582} \and
Abhijith Punnappurath$^{1*}$ \and\\
Mahmoud Afifi$^{1*}$\orcidlink{0000-0003-0125-4945} \and
Sangmin Han\inst{2}\orcidlink{0009-0001-6267-7290} \and
Alex Levinshtein$^1$ \and
Michael S. Brown$^1$
}

\authorrunning{D.~Kim et al.}

\vspace{-2mm}
\institute{
  $^1$AI Center - Toronto, Samsung Electronics \,
  $^2$Yonsei University \,
  $^*$Equal contribution
}

\maketitle

\begin{center}
    \vspace{-3mm}
    \footnotesize \textbf{Project Page:} \url{https://dy112.github.io/rawgen-page/}
\end{center}

\begin{center}
    \centering
    \captionsetup{type=figure} 
    \includegraphics[width=0.99\textwidth]{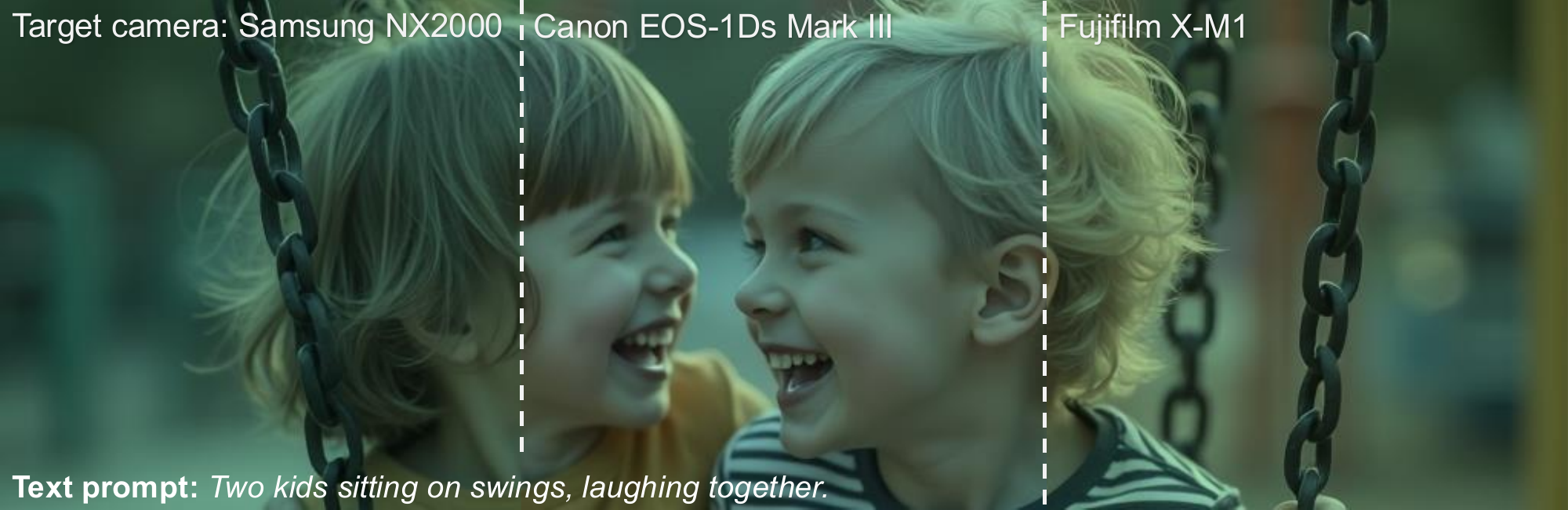}
    \vspace{-2mm}
    \captionof{figure}{We present RawGen, a diffusion-based method for generating realistic camera raw images. RawGen produces a latent representation of linear CIE XYZ images, conditioned on an sRGB image or a text prompt. The latent is decoded to CIE XYZ and mapped to arbitrary target camera raw spaces.}
    \label{fig:teaser}
    \vspace{-2mm}
\end{center}

\newcommand{\Etal}   {{\textit{et al.}}}

\newcommand{\jy}[1]{{\textbf{\textcolor{MidnightBlue}{[JY] }}\textcolor{MidnightBlue}{#1}}}

\newcommand{\change}[1]{{\color{red}#1}}

\newcommand{\cm}{\checkmark}

\definecolor{lightlightgray}{gray}{0.96}

\begin{abstract}
\vspace{-1mm}
Cameras capture scene-referred linear raw images, which are processed by onboard image signal processors (ISPs) into display-referred 8-bit sRGB outputs. Although raw data is more faithful for low-level vision tasks, collecting large-scale raw datasets remains a major bottleneck, as existing datasets are limited and tied to specific camera hardware. Generative models offer a promising way to address this scarcity; however, existing diffusion frameworks are designed to synthesize photo-finished sRGB images rather than physically consistent linear representations. This paper presents RawGen, to our knowledge the first diffusion-based framework enabling text-to-raw generation for arbitrary target cameras, alongside sRGB-to-raw inversion. RawGen leverages the generative priors of large-scale sRGB diffusion models to synthesize physically meaningful linear outputs, such as CIE XYZ or camera-specific raw representations, via specialized processing in latent and pixel spaces. To handle unknown and diverse ISP pipelines and photo-finishing effects in diffusion-model training data, we build a many-to-one inverse-ISP dataset where multiple sRGB renditions of the same scene generated using diverse ISP parameters are anchored to a common scene-referred target. Fine-tuning a conditional denoiser and specialized decoder on this dataset allows RawGen to obtain camera-centric linear reconstructions that effectively invert the rendering pipeline. We demonstrate RawGen's superior performance over traditional inverse-ISP methods that assume a fixed ISP. Furthermore, we show that augmenting training pipelines with RawGen’s scalable, text-driven synthetic data can benefit downstream low-level vision tasks.
\end{abstract}
\section{Introduction}
\label{sec:intro}

Camera sensors natively record scene-referred linear raw data. Raw images are subsequently processed by onboard image signal processors (ISPs) through a lossy rendering pipeline to produce display-referred outputs, typically in the 8-bit sRGB format. While raw captures preserve physically meaningful radiometric information, they are rarely available at scale and often tied to specific camera models. Collecting large raw datasets is costly and labor-intensive, and data must be re-captured whenever the sensor or the camera pipeline changes, limiting reproducibility and cross-device coverage. This scarcity remains a fundamental bottleneck for low-level vision and computational photography research that relies on linear, sensor-centric measurements.

Generative models offer a potential route to mitigate this limitation. Recent diffusion-based frameworks have achieved remarkable success in producing high-fidelity, semantically coherent sRGB images~\cite{song2019generative, ho2020denoising, song2020score, nichol2021improved, song2020denoising, dhariwal2021diffusion, ho2022classifier, dit, pernias2023wurstchen, karras2022elucidating, song2023consistency}. In particular, text-to-image (T2I) systems~\cite{rombach2022high, saharia2022photorealistic, podell2023sdxl, esser2024scaling, batifol2025flux, ramesh2021zero, ramesh2022hierarchical, betker2023improving, zhou2024transfusion, dai2023emu, zhang2023adding} enable strong semantic controllability via textual prompts, facilitating scalable and diverse image synthesis. However, these models operate primarily in the 8-bit sRGB domain, which is optimized for display and storage rather than preserving physical linearity. This representation typically includes nonlinear photo-finishing operations (e.g., gamma correction, tone mapping, and color styling) that distort the underlying radiometric structure of the scene.
Consequently, diffusion model outputs are not directly compatible with scene-referred linear signals, such as camera raw measurements or device-independent linear representations that can serve as canonical references.

A natural strategy is to reconstruct linear radiometric data from diffusion-generated sRGB images using inverse-ISP techniques~\cite{Xing2021invisp, afifi2021cie, reinders2025raw, berdan2025reraw}.
However, most existing inverse-ISP methods are trained on paired datasets generated under a fixed imaging pipeline, such as software ISPs (e.g., RawPy, DCRAW) or a specific in-camera ISP configuration. These methods therefore implicitly assume a single camera response and a known sequence of color and tone transformations. In contrast, diffusion models are trained on large-scale sRGB corpora that encompass diverse and heterogeneous photo-finishing styles originating from unknown ISPs and post-processing pipelines.

This discrepancy creates a fundamental domain gap. Conventional inverse-ISP models expect sRGB inputs produced by a fixed and known rendering pipeline, whereas diffusion-generated sRGB images entangle unknown, diverse, and often highly nonlinear photo-finishing effects. Directly applying standard inverse-ISP methods to diffusion model outputs fails to reliably recover a consistent scene-referred linear representation. Enabling physically meaningful linear signal synthesis from generative models therefore remains an open challenge.

In this paper, we introduce RawGen, a diffusion-based framework designed to bridge this gap. RawGen preserves the strong semantic priors and promptability of large-scale diffusion models while incorporating learned latent-level and pixel-level unprocessing to recover a canonical scene-referred linear representation from diverse sRGB inputs. Rather than assuming a fixed inverse mapping, RawGen is trained to suppress uncontrolled and heterogeneous photo-finishing effects implicitly embedded in diffusion-generated images, thereby enabling consistent and physically grounded linear synthesis.

The central idea of RawGen is to factor out ISP-induced variability and recover a shared linear anchor that remains consistent across multiple sRGB versions of the same underlying scene. To achieve this, we adopt a many-to-one reconstruction objective in which multiple photo-finished sRGB images of a single scene serve as inputs and are mapped to a common linear reference. This formulation explicitly encourages the recovered linear representation to remain invariant to differences in post-processing operations—such as tone mapping, gamma correction, and color styling—that are embedded in the sRGB inputs, while preserving the underlying scene structure and semantic content.

With this learned invariance, RawGen supports two complementary modes of generation. First, it converts an sRGB image into a canonical scene-referred linear representation that is robust to diverse post-processing styles. Second, it maps nonlinear image representations produced from text prompts by a pretrained diffusion prior into canonical linear representations. When a target camera raw space is required, the canonical linear output can be transformed into the camera-specific raw domain using standard color-space mappings and camera metadata, enabling camera-agnostic raw synthesis without retraining (\cref{fig:teaser}).

We demonstrate that RawGen consistently outperforms conventional inverse-ISP methods that rely on fixed imaging assumptions, yielding a more stable and scene-referred linear representation under diverse and uncontrolled photo-finishing variations. Moreover, RawGen enables scalable text-driven raw generation as a practical source of training data for downstream tasks such as illuminant estimation, neural ISP learning, and denoising, providing broad scene diversity without additional capture effort or scene asset design.
\vspace{-2mm}

\subsection*{Contribution}

In summary, our main contributions are as follows:
\begin{itemize}[leftmargin=10pt, label=\raisebox{0.25ex}{\tiny$\bullet$}]

    \item We propose RawGen, a diffusion-based camera-agnostic raw generation framework that preserves strong generative priors while unprocessing diverse sRGB inputs with unknown photo-finishing into a canonical scene-referred linear domain via latent- and pixel-level processing.
    \item We introduce a many-to-one linear reconstruction task and propose a method to construct a new type of dataset that maps multiple photo-finished sRGB observations to a single common linear reference, enabling robustness to heterogeneous and unknown ISP transformations.
    \item We develop, to our knowledge, the first unified framework enabling text-driven synthesis of physically meaningful linear and camera-specific raw data for arbitrary target cameras, and demonstrate that the resulting synthetic data alleviates data acquisition challenges in downstream low-level vision tasks.
\end{itemize}

\section{Related Works}
\label{sec:related_works}

\vspace{-4mm}
\paragraph{Forward \& Inverse Camera ISP Pipelines.}
Traditional camera hardware ISPs consist of carefully engineered signal-processing stages~\cite{delbracio2021mobile}, such as denoising, demosaicing, white balancing, color correction, tone mapping, and more, often tuned to produce a manufacturer-specific visual aesthetic. Because in-camera pipelines are proprietary and largely inaccessible, the rendered sRGB photographs available on the internet reflect diverse and unknown photo-finishing operations. Large-scale generative models trained on such data therefore inherit these uncontrolled rendering characteristics, producing outputs in the nonlinear sRGB domain with implicit and heterogeneous ISP effects. 

Recent work has explored learning-based ISPs that replace handcrafted modules with end-to-end neural mappings (e.g.,~\cite{ignatov2020replacing,lan,microisp,zhang2021learning,fourier}). These networks are trained to convert raw images, typically from a specific camera, to their display-referred outputs assuming a one-to-one relationship between the sRGB images and their originating ISPs.

Recovering linear sensor data from rendered images has long been studied for low-level vision. Early approaches to raw reconstruction relied on calibration-based techniques~\cite{debevec2008recovering,mitsunaga1999radiometric,grossberg2003determining,chakrabarti2014modeling,Chakrabarti2009empirical,kim2012new}. Later work, such as UPI~\cite{brooks2019unprocessing}, performed sRGB-to-raw reconstruction by sequentially inverting the ISP with non-learnable parametric operations. Nam et al.~\cite{nam2017modelling} proposed one of the earliest deep learning frameworks to jointly model the forward and inverse ISP mappings. Building on this idea, subsequent methods such as CIE-XYZ-Net~\cite{afifi2021cie}, CycleISP~\cite{zamir2020cycleisp}, InvISP~\cite{Xing2021invisp}, ParamISP~\cite{Kim_2024_CVPR}, and model-based ISP approaches~\cite{conde2022model} further advanced the field through convolutional networks and differentiable, model-driven architectures trained on paired raw-sRGB data.

Despite their differences, existing data-driven inverse ISP approaches largely assume a one-to-one relationship between rendered images and their underlying ISPs. This assumption is reasonable when training data is captured or synthesized under controlled pipelines but breaks down for diffusion-generated imagery, where rendering parameters are unknown and highly variable. Consequently, applying conventional inverse-ISP strategies to generative outputs cannot reliably recover a consistent scene-referred representation.

\paragraph{Diffusion Models for Low-Level Vision Tasks.}
Denoising diffusion models~\cite{ho2020denoising, song2020denoising, rombach2022high} are powerful generative models that synthesize images by reversing a gradual noising process. Their state-of-the-art performance has been demonstrated across diverse low-level vision tasks~\cite{luo2024flowdiffuser, saharia2022palette}, including super-resolution~\cite{saharia2022image}, image restoration~\cite{wang2022zero, kawar2022denoising}, and low-light enhancement~\cite{jiang2023low, wang2023exposurediffusion}. Most methods approach these tasks as conditional generation problems, guiding the diffusion process with auxiliary inputs like degraded images or semantic maps.

Recently, diffusion-based approaches have been explored for ISP-related tasks. Some works perform inverse-ISP by reconstructing raw from sRGB inputs~\cite{reinders2025raw}, while others learn forward ISP mappings (raw-to-sRGB)~\cite{chen2025rddm, ren2025ispdiffuser}. These methods, however, typically rely on synthetic training pairs generated via fixed one-to-one ISP logic (e.g., using RawPy). More recently, Yuan et al.~\cite{yuan2025generative} proposed controlling camera parameters (e.g., white balance, exposure) by guiding the diffusion process to maintain scene consistency. While effective for guided synthesis, this approach embeds parameter adjustment within the iterative generation, potentially requiring a new generative pass for each modification. For HDR reconstruction, diffusion-based works~\cite{wang2025lediff, bemana2025bracket} focus on fusing multiple exposures, generally outputting tone-mapped sRGB or display-oriented HDR formats.

Consequently, these approaches do not address the challenge of reconstructing camera-centric, physically linear representations (e.g., CIE XYZ or raw) from in-the-wild sRGB inputs. Furthermore, they do not explore text-driven synthesis of raw data. 

Such an approach that generates raw data would decouple the initial generation from subsequent adjustments, thereby enabling flexible and efficient post-processing in standard dedicated software.

To handle unknown ISP pipelines, we adopt a many-to-one training paradigm that anchors diverse photo-finished observations to a shared linear target, enabling physically consistent raw generation at scale. RawGen is the first framework capable of synthesizing physically meaningful raw images directly from open-world text prompts without assuming a fixed imaging pipeline.
\section{Method}
\label{sec:method}

RawGen synthesizes physically meaningful linear data by repurposing pretrained sRGB diffusion models. As discussed in \cref{sec:related_works}, the same raw capture can yield substantially different sRGB renderings under different ISP and photo-finishing settings, and large-scale diffusion training corpora entangle these unknown, heterogeneous effects. Rather than assuming a fixed inverse mapping as in conventional inverse-ISP methods, RawGen learns to suppress this uncontrolled variability and recover a canonical, scene-referred linear representation.

We adopt CIE XYZ as the canonical space. It is linear and device-independent, and it relates to camera-specific raw spaces through standard color transforms, enabling camera-agnostic synthesis without retraining.

\begin{figure*}[!t]
\centering
\includegraphics[width=\textwidth]{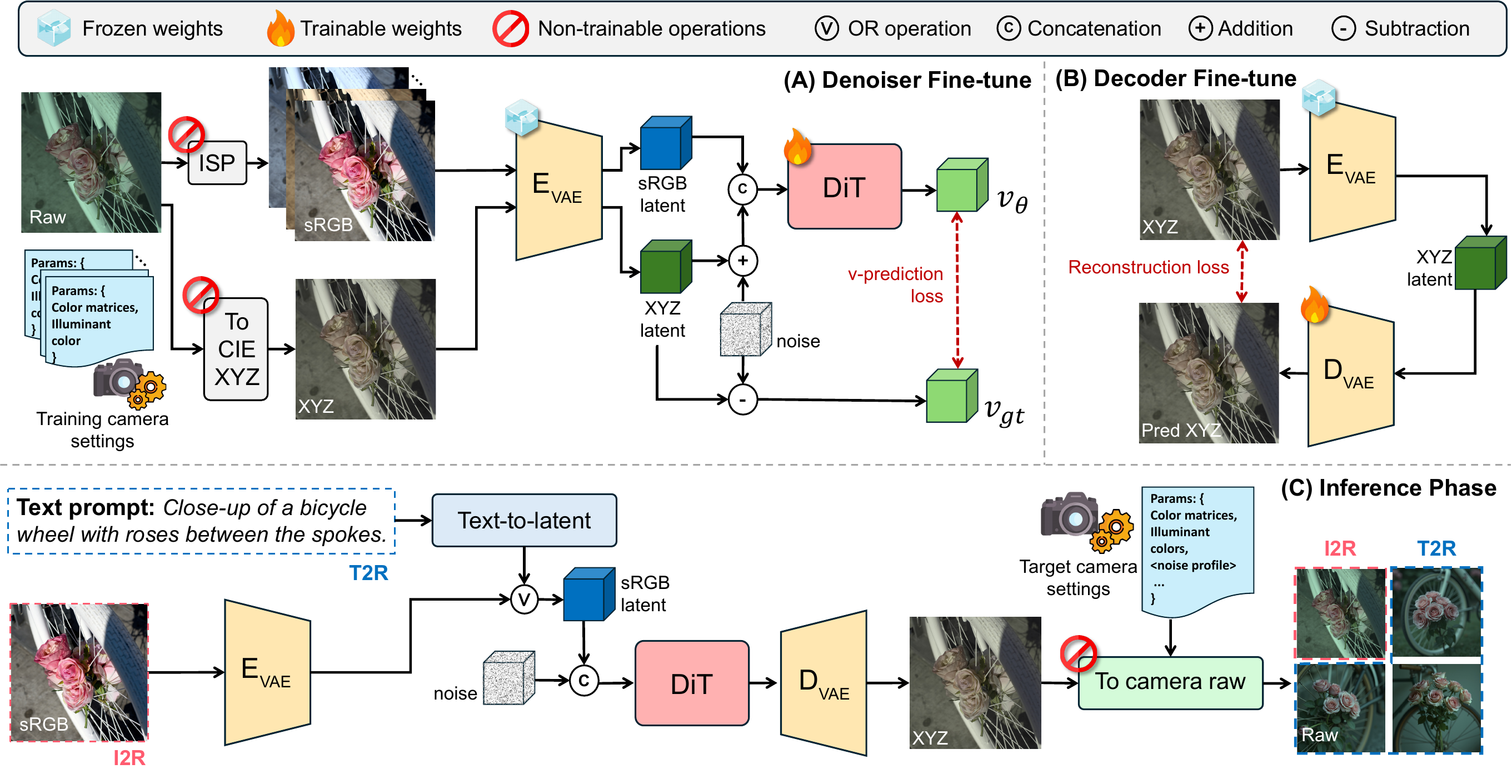}
\vspace{-4mm}
\caption{Overview of the RawGen framework. During training, raw images are converted to CIE XYZ and sRGB representations to (A) fine-tune the DiT to denoise CIE XYZ latents conditioned on an sRGB image, and (B) fine-tune the VAE decoder to reconstruct CIE XYZ images. During inference (C), either an sRGB image (image-to-raw, I2R) or a text prompt (text-to-raw, T2R) is used to condition the DiT to generate a CIE XYZ latent, which is decoded to obtain a CIE XYZ image and subsequently mapped to the target camera's raw space using its calibration metadata, which can be easily acquired from a single DNG file of target camera.\vspace{-4mm}}
\label{fig:rawgen_overview}
\end{figure*}

As shown in \cref{fig:rawgen_overview}, our method builds on a pretrained rectified-flow DiT~\cite{dit} with native image conditioning capability. We first describe the many-to-one training data construction (\cref{sec:data_generation}) that underlies our training stages, then present the two fine-tuning stages: denoiser fine-tuning (\cref{sec:denoiser_finetune}) and decoder fine-tuning (\cref{sec:decoder_finetune}), corresponding to panels~(A) and~(B) of \cref{fig:rawgen_overview}, and finally detail inference for image-to-raw and text-to-raw generation (\cref{sec:inference}), corresponding to \cref{fig:rawgen_overview}(C).

\subsection{Many-to-One Training Data Construction}
\label{sec:data_generation}

To invert diverse, unknown photo-finishing effects into a single linear target, we first construct paired data that links multiple sRGB renditions of the same scene to one common anchor. Starting from a raw image $I_{\mathrm{raw}}$, we derive a linear CIE XYZ image $I_{\mathrm{XYZ}}$ by applying per-scene white balancing and a camera-to-XYZ color conversion matrix, yielding an illumination-neutral, scene-referred representation that precedes any photo-finishing. We treat $I_{\mathrm{XYZ}}$ as the \emph{anchor}, shared across all sRGB variants of the same scene.

Using this anchor, we generate $N$ sRGB variants $\{I_{\mathrm{sRGB}}^{(n)}\}_{n=1}^{N}$ by sampling diverse photo-finishing parameter vectors $P^{(n)}$, which control global color tone, contrast, and tone mapping. Concretely, we randomize three ISP parameter groups: (i) white balance, by perturbing red and blue gains around the DNG \texttt{AsShotNeutral} initialization; (ii) tone mapping, by varying the per-channel tone-curve shape; and (iii) contrast, via a global gain about a mid-gray pivot in sRGB. Together, these perturbations yield diverse photo-finishing styles while preserving the same scene content. By pairing multiple sRGB variants with a single XYZ anchor, this strategy approximates the heterogeneous ISP behavior found in real-world sRGB images, encouraging invariance to photo-finishing while preserving scene content. Each image is encoded by the frozen VAE encoder $E_{\text{VAE}}$:
\begin{equation}
\label{eq:vae-enc}
z_{\mathrm{sRGB}}^{(n)} = E_{\text{VAE}}\!\left(I_{\mathrm{sRGB}}^{(n)}\right), \quad
z_{\mathrm{XYZ}} = E_{\text{VAE}}\!\left(I_{\mathrm{XYZ}}\right).
\end{equation}
The resulting latent pairs $\{(z_{\mathrm{sRGB}}^{(n)},\, z_{\mathrm{XYZ}})\}_{n=1}^{N}$ are cached for both training stages.
More details of the ISP pipeline and photo-finishing parameter sampling are provided in the supplementary material.

\subsection{Denoiser Fine-Tuning}
\label{sec:denoiser_finetune}

The first training stage, shown in \cref{fig:rawgen_overview}(A), fine-tunes the pretrained DiT to map sRGB-conditioned inputs to the linear XYZ anchor latent.

\paragraph{Objective.}
Given the anchor latent $z_{\mathrm{XYZ}}$ from \cref{eq:vae-enc}, we randomly sample one sRGB variant latent $z_{\mathrm{sRGB}}^{(n)}$ under the many-to-one setting. We corrupt the anchor latent by interpolating with Gaussian noise $\epsilon \sim \mathcal{N}(0, \mathbf{I})$:
\begin{equation}
\label{eq:forward}
z_t = (1 - t)\, z_{\mathrm{XYZ}} + t\, \epsilon, \quad t \in [0, 1],
\end{equation}
and tune the DiT parameterized by $\theta$ to predict the rectified-flow velocity target (also referred to as a $v$-prediction target)
$v_{\mathrm{gt}} = \epsilon - z_{\mathrm{XYZ}}$:
\begin{equation}
\label{eq:ldenoise}
\mathcal{L}_{\text{denoise}}
= \mathbb{E}_{n,\,t,\,\epsilon}
\left\|
v_{\mathrm{gt}} - v_{\theta}\!\left(z_t,\, t;\, z_{\mathrm{sRGB}}^{(n)}\right)
\right\|_2^2.
\end{equation}
By tuning with randomly sampled variants, the denoiser learns to suppress ISP-induced variability in the sRGB condition and consistently recover the shared latent representation of the linear anchor XYZ image.

\paragraph{Conditioning.}
We leverage the image conditioning mechanism of the pretrained DiT. The sRGB context latent $z_{\mathrm{sRGB}}^{(n)}$ and the noisy target latent $z_t$ are patchified into token sequences and concatenated along the sequence dimension, and the combined sequence is processed jointly through the DiT transformer blocks. Only the target portion of the output contributes to the loss. We fine-tune the model using LoRA adapters~\cite{hu2022lora} on attention projection layers, while keeping the pretrained backbone weights frozen.

\subsection{Decoder Fine-Tuning}
\label{sec:decoder_finetune}

The second training stage, shown in \cref{fig:rawgen_overview}(B), fine-tunes the VAE decoder to reconstruct linear XYZ images from XYZ-domain latents. Since the pretrained decoder is optimized for sRGB, directly decoding XYZ latents may yield degraded outputs. We fine-tune $D_{\text{VAE}}$ using the ground-truth anchor latent $z_{\mathrm{XYZ}}$ from \cref{eq:vae-enc}, minimizing an $\ell_1$ loss:
\begin{equation}
\label{eq:recon}
\widehat{I}_{\mathrm{XYZ}} = D_{\text{VAE}}\!\left(z_{\mathrm{XYZ}}\right),
\;
\mathcal{L}_{\text{recon}} = \left\|\widehat{I}_{\mathrm{XYZ}} - I_{\mathrm{XYZ}}\right\|_1.
\end{equation}
This retargets the VAE decoder from sRGB to linear XYZ while preserving the spatial representations learned during pretraining.

\subsection{Inference: Image-to-Raw and Text-to-Raw}
\label{sec:inference}

As illustrated in \cref{fig:rawgen_overview}(C), RawGen uses a unified inference pipeline that first synthesizes a scene-referred CIE XYZ image and then renders it into a target camera's linear raw space. The only difference between image-to-raw (I2R) and text-to-raw (T2R) lies in how the sRGB conditioning latent is obtained within the same pretrained backbone that shares the core DiT and the VAE encoder/decoder. Specifically, I2R encodes an input sRGB image using the frozen VAE encoder, whereas T2R follows the conventional diffusion model's standard text-conditioned generation route and directly takes the intermediate latent produced from the text prompt prior to VAE decoding (i.e., the latent that would otherwise be decoded to an sRGB image). In both cases, this sRGB latent conditions the same DiT to generate an XYZ latent, which is decoded with our fine-tuned VAE decoder and finally mapped deterministically to the target camera's raw space using its calibration metadata under a chosen illuminant. In our implementation, we instantiate the backbone with FLUX.1 and reuse its native text-to-latent pathway for T2R conditioning.

\paragraph{Image-to-Raw (I2R).}
Given an sRGB image $I_{\mathrm{sRGB}}$, we obtain the conditioning latent
$z_{\mathrm{sRGB}} = E_{\text{VAE}}(I_{\mathrm{sRGB}})$.
We initialize the target tokens with noise and solve the reverse ODE conditioned on $z_{\mathrm{sRGB}}$:
\begin{equation}
\label{eq:sampling}
\hat{z}_{\mathrm{XYZ}} = z_1 + \int_{1}^{0} v_\theta\!\left(z_t,\, t;\, z_{\mathrm{sRGB}}\right) \mathrm{d}t,
\end{equation}
approximated using Euler steps. We retain only the target tokens, reshape them into the spatial latent $\hat{z}_{\mathrm{XYZ}}$, and decode
$\widehat{I}_{\mathrm{XYZ}} = D_{\text{VAE}}(\hat{z}_{\mathrm{XYZ}})$.
The decoded XYZ image is then converted to the target camera raw space using the mapping described below.

\paragraph{Text-to-Raw (T2R).}
Given a text prompt, we first obtain an sRGB conditioning latent $z_{\mathrm{sRGB}}$ using the pretrained text-to-latent pathway of the base model. We then run the same conditional generation and decoding steps as in I2R to produce $\hat{z}_{\mathrm{XYZ}}$ and $\widehat{I}_{\mathrm{XYZ}}$, and apply the same XYZ-to-camera mapping to obtain camera-specific linear raw outputs.

\paragraph{CIE XYZ to Camera Raw.}
To convert $\widehat{I}_{\mathrm{XYZ}}$ into a target camera's linear raw-RGB space, we apply calibration transforms derived from the target camera's DNG metadata together with a chosen illuminant parameterized by correlated color temperature (CCT). In practice, we (i) interpolate camera calibration matrices based on the selected CCT, (ii) map $\widehat{I}_{\mathrm{XYZ}}$ to a white-balanced camera RGB using the interpolated forward model, and (iii) apply an illuminant gain in camera RGB space to obtain the illuminated camera RGB, which we treat as the target camera's linear raw-RGB representation. This mapping is deterministic and non-trainable, enabling re-rendering of the same scene-referred output across cameras without retraining. Full details of matrix interpolation and illuminant sampling are provided in the supplementary material.

\paragraph{Optional Noise and CFA.}
If required, we synthesize sensor noise using a heteroscedastic noise model to better match the noise characteristics of real raw camera images; we refer the reader to the supplementary material for details. While more advanced noise synthesis models (e.g., \cite{abdelhamed2019noise}) can produce more realistic noise characteristics, detailed noise modeling is beyond the scope of this paper.
Finally, if required by the downstream task, the resulting linear RGB raw image can be re-mosaiced according to a specified color filter array (CFA) pattern to produce a single-channel mosaiced raw image.
\section{Experiments}
\subsection{Training Details and Experiment Overview}
\paragraph{Training Details.}
We train RawGen using the MIT-Adobe FiveK~\cite{fivek} and RAISE~\cite{dang2015raise} datasets. Both datasets provide large-scale raw DNG images covering diverse scenes and subjects, including urban and natural environments, as well as animals and human portraits.
To obtain CIE XYZ anchor targets, we parse the \texttt{AsShotNeutral} and \texttt{ForwardMatrix} tags from each DNG file. We then apply the software ISP~\cite{Seo2023graphics2raw} with randomized photo-finishing parameters to render diverse sRGB variants from the same anchor (\cref{sec:data_generation}).
For the DiT backbone, we adopt FLUX.1-Kontext~\cite{batifol2025flux}, which natively supports image-context conditioning. We fine-tune the DiT model via LoRA with rank $r{=}64$ and scaling $\alpha{=}64$.

\paragraph{Experiment Overview.}
As described in \cref{sec:method}, once trained, RawGen synthesizes scene-referred linear representations under two conditioning modes: (i) an input sRGB image (I2R) and (ii) a text prompt (T2R). We evaluate RawGen from three perspectives. First, we assess its many-to-one reconstruction capability by recovering a canonical linear XYZ representation from sRGB inputs with diverse and unknown photo-finishing (\cref{sec:n_to_one}). Second, we evaluate device-specific raw synthesis by mapping canonical outputs to target camera domains and examining their distribution alignment with real raw data (\cref{sec:raw_synth}). Third, we study practical applications enabled by RawGen, including scalable T2R-based data generation for downstream low-level vision tasks and raw-domain image editing. (\cref{ssec:app}).
Together, these experiments demonstrate that RawGen enables physically grounded linear synthesis while supporting camera-aware and scalable raw generation.

\subsection{Evaluation of Many-to-One sRGB-to-XYZ Invertability}
\label{sec:n_to_one}
RawGen projects sRGB inputs containing unknown and diverse photo-finishing effects into a shared canonical scene-referred XYZ domain, suppressing photo-finishing variability while preserving scene content. To evaluate this many-to-one inverse-ISP capability, we conduct two complementary analyses using the Image-to-Raw (I2R) inference pathway (\cref{fig:rawgen_overview}C): (1) expert-retouched real variations and (2) text-guided synthetic variations. We compare RawGen against CIE XYZ Net~\cite{afifi2021cie}, InvISP~\cite{Xing2021invisp}, and Raw-Diffusion~\cite{reinders2025raw}.
Since these baselines typically assume a one-to-one correspondence between an sRGB input and its reconstruction target (e.g., raw or XYZ), we train all methods to invert the ISP induced by the \emph{default} configuration of our software ISP, using the corresponding (default sRGB, anchor XYZ) pair as supervision.

\paragraph{Expert-Retouched Real Variations.}
For this experiment, we use the MIT-Adobe FiveK dataset~\cite{fivek}, where each scene is edited by five experts (A--E) with distinct aesthetic preferences.
For each scene, the five expert-retouched sRGB images are fed into RawGen, and reconstruction accuracy is measured against the reference scene-referred XYZ target using PSNR and SSIM. All evaluations are conducted on test splits unseen during training for both RawGen and the compared methods. As a baseline, we evaluate $\text{RawGen}_{\text{one-to-one}}$, which uses the same architecture as RawGen but is trained with strict one-to-one supervision on (default sRGB, anchor XYZ) pairs rather than our many-to-one training scheme.

\begin{table}[!t]
  \caption{Quantitative comparison of sRGB-to-XYZ reconstruction across five photo-finishing styles on the MIT-Adobe FiveK dataset~\cite{fivek}. We report results for ablated variant of RawGen. Best results are highlighted in \textbf{\colorbox{best}{yellow}}.\label{tab:expert_psnr_ssim}}
  \centering
\resizebox{\textwidth}{!}{
\begin{tabular}{|l|cc|cc|cc|cc|cc|}
\hline
 & \multicolumn{2}{c|}{\cellcolor{shade1}Expert A} & \multicolumn{2}{c|}{\cellcolor{shade2}Expert B} & \multicolumn{2}{c|}{\cellcolor{shade3}Expert C} & \multicolumn{2}{c|}{\cellcolor{shade4}Expert D} & \multicolumn{2}{c|}{\cellcolor{shade5}Expert E} \\ \cline{2-11} 
\multirow{-2}{*}{Method} & \cellcolor{shade1-1}PSNR & \cellcolor{shade1-1}SSIM & \cellcolor{shade2-1}PSNR & \cellcolor{shade2-1}SSIM & \cellcolor{shade3-1}PSNR & \cellcolor{shade3-1}SSIM & \cellcolor{shade4-1} PSNR & \cellcolor{shade4-1} SSIM & \cellcolor{shade5-1} PSNR & \cellcolor{shade5-1} SSIM \\ \hline
CIE XYZ Net \cite{afifi2021cie} & 19.60 & 0.7861 & 21.04 & 0.8431 & 19.49 & 0.7795 & 19.44 & 0.8058 & 18.64 & 0.7918 \\
InvISP \cite{Xing2021invisp} & 16.04 & 0.6854 & 16.04 & 0.6854 & 14.92 & 0.6323 & 14.24 & 0.6802 & 13.30 & 0.6473 \\
Raw-Diffusion \cite{reinders2025raw} & 19.30 & 0.7832 & 20.66 & 0.8397 & 18.79 & 0.7705 & 18.69 & 0.7966 & 17.49 & 0.7751 \\\cdashline{1-11}

RawGen$_\text{one-to-one}$ &
19.57 &
0.7782 &
21.21 &
0.8349 &
19.48 &
0.7741 &
18.74 &
0.7929 &
18.07 &
0.7839 \\\hline



\textbf{RawGen (Ours)} &
\textbf{\cellcolor{best} 23.20} &
\textbf{\cellcolor{best} 0.8432} &
\textbf{\cellcolor{best} 24.35}  & 
\textbf{\cellcolor{best} 0.8581} & 
\textbf{\cellcolor{best} 23.37} & 
\textbf{\cellcolor{best} 0.8387} & 
\textbf{\cellcolor{best} 23.51} & 
\textbf{\cellcolor{best} 0.8531}  & 
\textbf{\cellcolor{best} 23.89} & 
\textbf{\cellcolor{best} 0.8500} \\\hline


\end{tabular}
}
\vspace{-2mm}
\end{table}
\begin{figure}[!t]
  \centering
  \includegraphics[width=0.98\columnwidth]{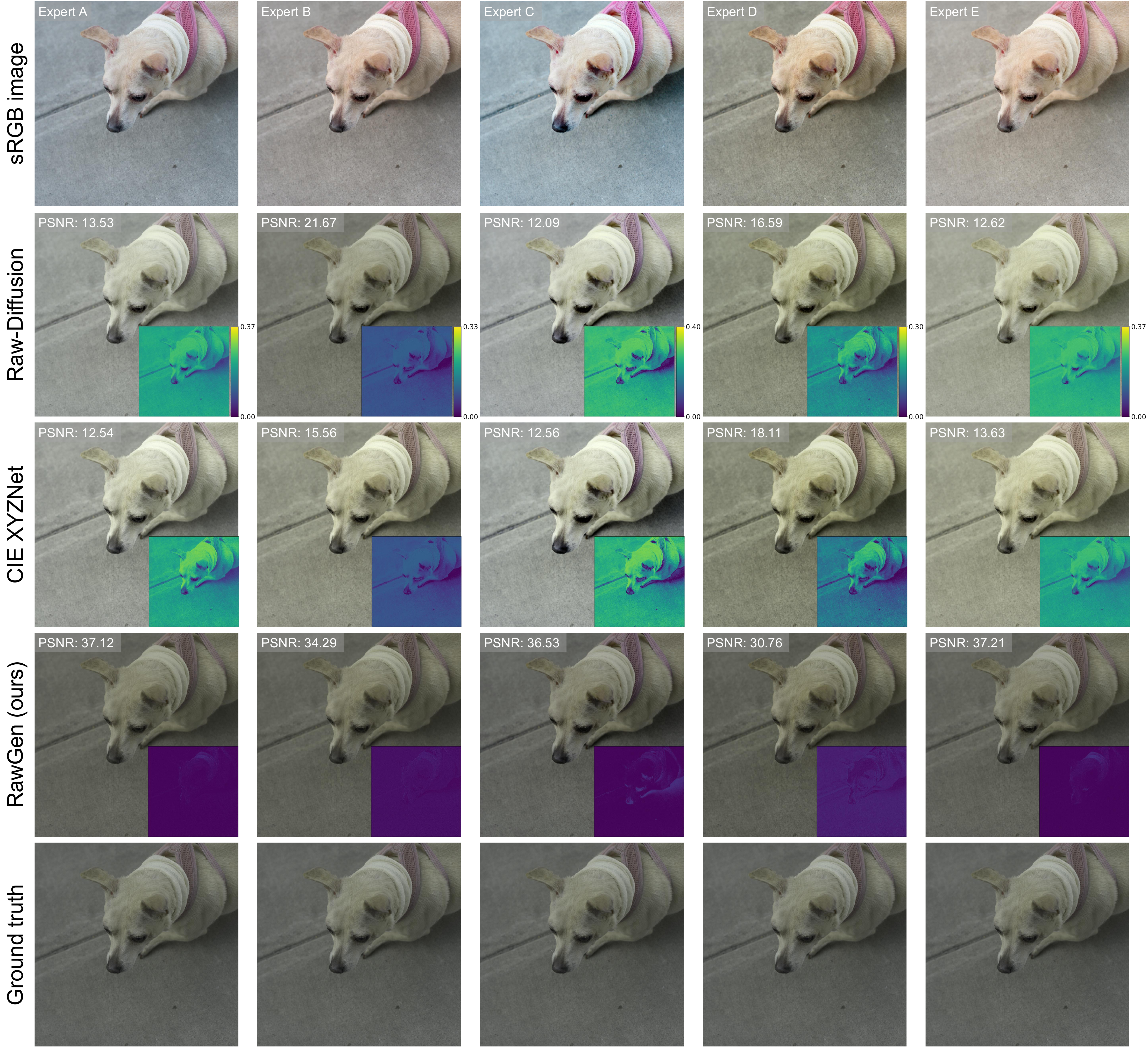}
  \vspace{-2mm}
\caption{CIE XYZ reconstruction results.
Shown is an sRGB input image rendered with different rendering preferences (Expert A--E) and the corresponding CIE XYZ reconstructions produced by Raw-Diffusion \cite{reinders2025raw}, CIE XYZ Net \cite{afifi2021cie}, and our RawGen. The last row shows the ground-truth CIE XYZ image.}
  \vspace{-2mm}
  \label{fig:xyz-qualitative}
\end{figure}

As shown in \cref{tab:expert_psnr_ssim}, RawGen consistently generalizes to unseen photo-finishing styles and more accurately recovers scene-referred linear XYZ representations than prior methods. The performance drop of $\text{RawGen}_{\text{one-to-one}}$ further highlights the importance of domain-level many-to-one supervision. Qualitative reconstructions and error maps in \cref{fig:xyz-qualitative} further corroborate RawGen’s robustness under diverse and previously unseen edits.

\begin{figure}[!t]
  \centering
  \begin{minipage}[t]{0.48\linewidth}
    \captionof{table}{Latent space compactness of sRGB-to-XYZ conversion methods. 100 color-graded sRGB variants per prompt are converted to XYZ and encoded into the VAE latent space; lower mean L2 distance to the centroid indicates more compact clusters.\label{tab:tsne_dist}}
    \vspace{1mm}
    \centering
    \resizebox{\linewidth}{!}{
    \setlength{\tabcolsep}{5pt}
    \begin{tabular}{
    |p{0.43\linewidth}| 
    >{\centering\arraybackslash}p{0.19\linewidth}
    >{\centering\arraybackslash}p{0.19\linewidth}
    >{\centering\arraybackslash}p{0.19\linewidth}|}
    \hline
    \multirow{2}{*}{Method} & \multicolumn{3}{c|}{Mean distance to centroid $\downarrow$} \\
                              & PCA    & t-SNE & UMAP  \\ \hline
    \tikz\draw[fill={rgb,255:red,106;green,27;blue,154},draw={rgb,255:red,106;green,27;blue,154}](0,0)circle(.45ex);\; sRGB         & 286.8  & 27.65 & 2.099 \\ \cdashline{1-4}
    \tikz\draw[fill={rgb,255:red,21;green,101;blue,192},draw={rgb,255:red,21;green,101;blue,192}](0,0)circle(.45ex);\; InvISP       & 265.4  & 24.33 & 1.913 \\
    \tikz\draw[fill={rgb,255:red,249;green,168;blue,37},draw={rgb,255:red,249;green,168;blue,37}](0,0)circle(.45ex);\; XYZNet       & 256.1  & 25.52 & 2.014 \\
    \tikz\draw[fill={rgb,255:red,198;green,40;blue,40},draw={rgb,255:red,198;green,40;blue,40}](0,0)circle(.45ex);\; RAW-Diffusion & 318.2  & 19.58 & 2.557 \\ \cdashline{1-4}
    \tikz\draw[fill={rgb,255:red,46;green,125;blue,50},draw={rgb,255:red,46;green,125;blue,50}](0,0)circle(.45ex);\; RawGen        & \textbf{\cellcolor{best}160.0} & \textbf{\cellcolor{best}10.69} & \textbf{\cellcolor{best}1.067} \\ \hline
    \end{tabular}}
  \end{minipage}
  \hfill
  \begin{minipage}[t]{0.48\linewidth}
    \vspace{11pt}
    \centering
    \includegraphics[width=0.9\linewidth]{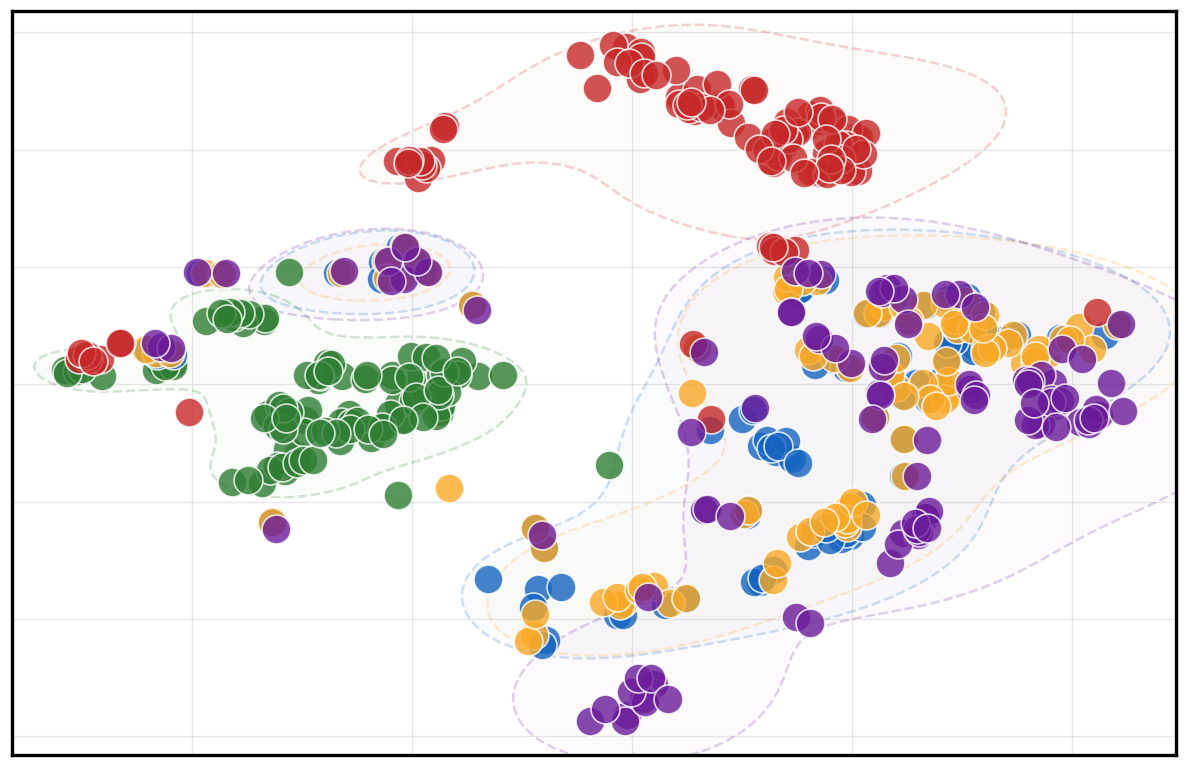}
    \vspace{-10pt}
    \captionof{figure}{t-SNE visualization of VAE latent spaces for each sRGB-to-XYZ
    conversion method. Dashed contours indicate the estimated density region per method.
    RawGen produces the most compact cluster.\label{fig:tsne_vis}}
  \end{minipage}
\end{figure}

\paragraph{Text-Augmented Synthetic Variations.}
To further examine many-to-one inverse-ISP capability, we construct a larger and more diverse set of sRGB variants using text-guided generation. Unlike the real-variant experiment above, which relies on five expert retouchings per scene, this setting synthesizes controlled color variations for a broader evaluation of linear invertibility.

We use the pre-trained FLUX.1-Kontext~\cite{batifol2025flux} in a two-stage process. First, an anchor sRGB image is generated from a prompt (e.g., ``\texttt{capture narrow alley after rainfall}''). The model is then conditioned on this anchor together with 100 color-grading prompts (e.g., ``\texttt{warm cast:}'', ``\texttt{cool cast:}'', ``\texttt{teal-and-orange grade:}''), producing 100 variants. Leveraging the image-conditioning feature of FLUX.1-Kontext, scene structure and texture are preserved, while the text prompts induce diverse global color shifts. These variants are converted to XYZ by each evaluated method, with RawGen operating via the Image-to-Raw inference pathway (\cref{fig:rawgen_overview}C).

To assess many-to-one consistency, the reconstructed outputs are encoded into the VAE latent space. The latent embeddings are projected using PCA~\cite{Pearson_1901_PCA}, t-SNE~\cite{Maaten_2008_JMLR}, and UMAP~\cite{McInnes_2018_UMAP}. PCA captures dominant linear variance, while t-SNE and UMAP preserve local neighborhood structure under nonlinear embeddings. Within each projected space, we compute the mean L2 distance to the centroid across variants. Lower distances indicate stronger suppression of photo-finishing variability and greater convergence toward a canonical representation.
\cref{tab:tsne_dist} reports the quantitative results, showing that RawGen produces more compact latent clusters than competing methods under PCA, t-SNE, and UMAP projections, whereas other approaches yield more dispersed embeddings, indicating weaker suppression of photo-finishing variability. The t-SNE visualization in \cref{fig:tsne_vis} further confirms that RawGen forms a tight and well-separated cluster with reduced overlap. This result aligns with the many-to-one objective, which encourages diverse sRGB variants of the same scene to converge toward a canonical XYZ representation and promotes invariance to photo-finishing effects.

\begin{figure}[!t]
  \centering
  \includegraphics[width=\columnwidth]{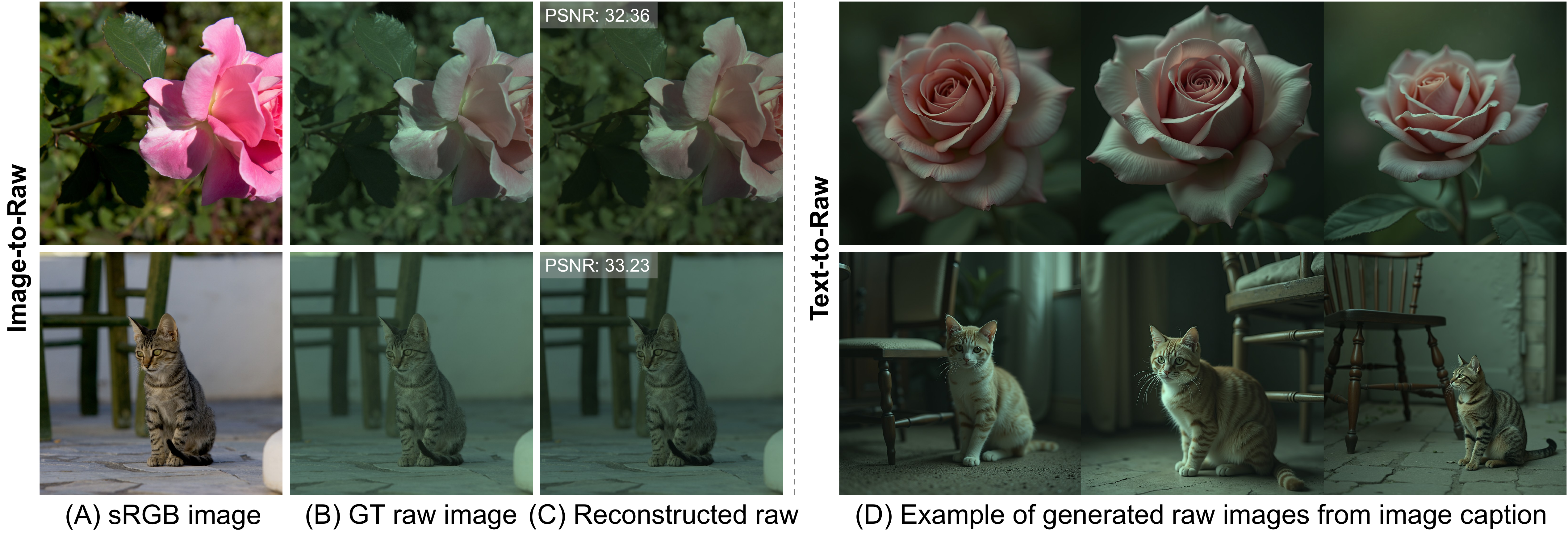}
\caption{
Raw reconstruction and generation results.
Shown are the input sRGB images (A) and corresponding ground-truth raw images (B). Our reconstructed raw results are shown in (C). Additional generated raw images of the same semantic scene are shown in (D), where InstructBLIP-2 \cite{dai2023instructblip} is used to generate text descriptions from image (A).}
  \label{fig:qual-results}
\end{figure}

\begin{figure}[!t]
  \centering
    \includegraphics[
      width=\columnwidth,
      trim=0 0 0 1.8cm,
      clip
  ]{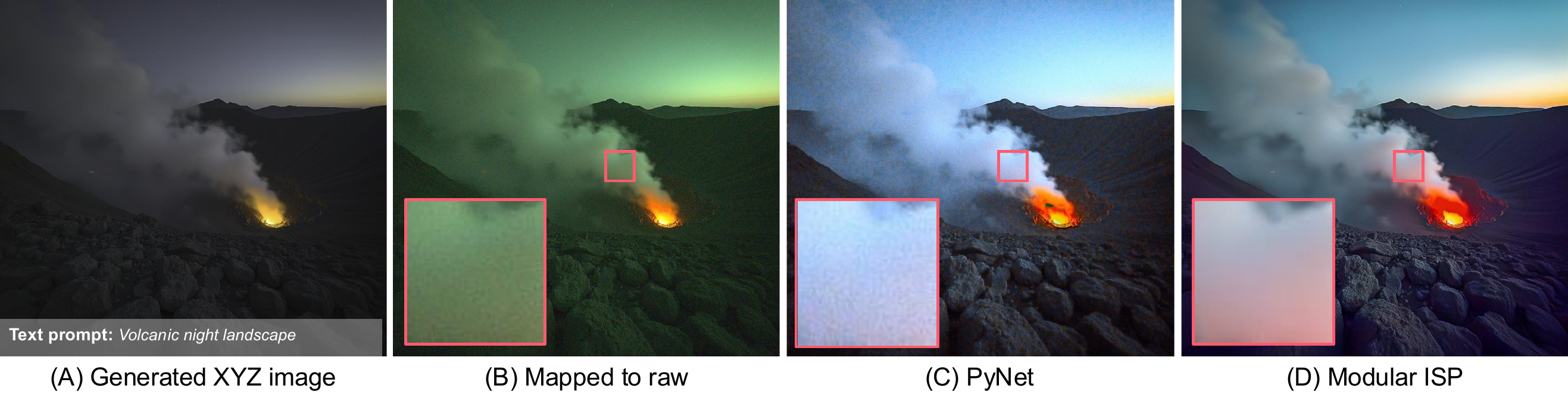}
\caption{Rendering results of pre-trained neural ISPs on real data from the Samsung S24 main camera (S24 dataset~\cite{afifi2025time}). (A) CIE XYZ image generated by our RawGen. (B) Image mapped to the S24 main camera raw space with synthetic noise. (C) Result of PyNet~\cite{ignatov2020replacing}. (D) Result of Modular Neural ISP~\cite{afifi2025modular}.\vspace{-6mm}}
  \label{fig:pre-trained-neural-isp-main}
\end{figure}

\subsection{Device-Specific Raw Synthesis}
\label{sec:raw_synth}
RawGen produces canonical linear XYZ representations independent of any specific camera. When a target device is specified, these outputs are mapped to the corresponding raw domain using device-specific color correction matrices (e.g., \texttt{ForwardMatrix}) and metadata, with heteroscedastic Gaussian noise injected to approximate sensor measurements. This decoupled design enables camera-agnostic linear generation and device-specific raw synthesis without retraining. \Cref{fig:qual-results} presents examples, including reconstruction from sRGB inputs with unknown edits (A–C) and multiple raw samples generated from text prompts via InstructBLIP-2~\cite{dai2023instructblip} (D), supporting scalable raw-domain augmentation.

We further assess distribution alignment between RawGen-synthesized and real-world raw images. Specifically, we evaluate neural ISP models trained on real raw data from the Samsung Galaxy S24 main camera (S24 dataset~\cite{afifi2025time}) on RawGen-generated raw images mapped to the same device domain, without retraining. As shown in \Cref{fig:pre-trained-neural-isp-main}, both ISP models produce plausible sRGB renderings. Notably, among the two evaluated ISPs, Modular Neural ISP includes an explicit denoising module and effectively suppresses the injected synthetic noise, indicating strong distribution alignment with real device-specific raw statistics.

\subsection{Application of RawGen}
\label{ssec:app}

\paragraph{Scalable Raw Synthesis for Downstream Tasks.}
Various low-level vision studies that rely on raw images have faced difficulties in data acquisition due to device-specific domain shifts, and collecting raw data at scale has required substantial human labor and extensive capture time. RawGen’s Text-to-Raw (T2R) synthesis flow enables scalable generation of diverse device-specific raw images without physical capture, making large-scale augmentation practical for low-level vision tasks. Having established the realism of these synthesized raw samples (\cref{sec:raw_synth}), we evaluate whether they can effectively support downstream tasks that require camera-specific raw data.

We evaluate impact of the scalability on three tasks: illuminant estimation, neural ISP learning, and raw-domain denoising, all requiring device-specific raw data. For controlled comparison, we follow the same datasets, training procedures, and evaluation protocols as Graphics2RAW~\cite{Seo2023graphics2raw}. For each task, we synthesize 3K samples. These synthetic samples are used only for training, and performance is measured on the corresponding real-world test splits without modification, isolating the effect of the synthetic data source while enabling direct comparison with prior methods.

The scalability advantage is particularly salient for illuminant estimation, which requires training data for nine cameras. In the real NUS-8 dataset, each camera provides only 217 images on average, and prior graphics-based raw synthesis~\cite{Seo2023graphics2raw} is further constrained by limited assets, yielding 125 synthetic raw images for training. In contrast, RawGen can generate diverse, camera-specific raw images directly from text prompts at scale, substantially expanding training coverage without additional capture effort or scene asset design.
Detailed experimental settings are provided in the supplementary material.

\begin{table}[!t]
  \caption{Quantitative comparison of downstream tasks: illuminant estimation (reported as angular error statistics), neural ISP learning, and denoising at two ISO levels. Each row corresponds to a model trained using data generated by the method listed. Models trained on real data are shown in the bottom row as a reference. The best results of models trained on \textit{non-real} raw data are highlighted in \textbf{\colorbox{best}{yellow}}.\label{tab:downstream_tasks}}
  \vspace{-2mm}
  \centering
\resizebox{0.96\textwidth}{!}{
\begin{tabular}{|l|ccc|ccc|cccc|}
\hline
 & \multicolumn{3}{c|}{\cellcolor{shade1}} & \multicolumn{3}{c|}{\cellcolor{shade2}} & \multicolumn{4}{c|}{\cellcolor{shade3}Denoising} \\
 & \multicolumn{3}{c|}{\cellcolor{shade1}\multirow{-2}{*}{Illuminant estimation}} & \multicolumn{3}{c|}{\multirow{-2}{*}{\cellcolor{shade2}Neural ISP}} & \multicolumn{2}{c}{\cellcolor{shade3-1}ISO 1600} & \multicolumn{2}{c|}{\cellcolor{shade3-2}ISO 3200} \\ \cline{2-4}
\multirow{-3}{*}{Method} & \cellcolor{shade1-1}Mean & \cellcolor{shade1-1}Median & \cellcolor{shade1-1} Worst 25\% & \cellcolor{shade2-1}PSNR & \cellcolor{shade2-1}SSIM & \cellcolor{shade2-1}$\Delta E$ & \cellcolor{shade3-1-1}PSNR & \cellcolor{shade3-1-1}SSIM & \cellcolor{shade3-2-1}PSNR & \cellcolor{shade3-2-1}SSIM \\ \hline
EnlightenGAN~\cite{jiang2021enlightengan} & 7.01 & 6.82 & 11.07 & 35.58 & 0.965 & 3.137 & 48.82 & 0.991 & 47.25 & 0.988 \\
UPI~\cite{brooks2019unprocessing} & 6.26 & 5.89 & 10.33 & 36.43 & 0.966 & 2.907 & 49.05 & 0.990 & 47.51 & 0.988 \\
Graphics2RAW~\cite{Seo2023graphics2raw} & 4.21 & 3.38 & 8.57 & 38.10 & \textbf{\cellcolor{best}0.974} & 2.301 & 49.37 & 0.991 & 48.16 & 0.989 \\\cdashline{1-11}
RawGen & \textbf{\cellcolor{best}3.14} & \textbf{\cellcolor{best}2.11} & \textbf{\cellcolor{best}7.37} & \textbf{\cellcolor{best}38.42} & 0.970 & \textbf{\cellcolor{best}2.183} & \textbf{\cellcolor{best}50.63} & \textbf{\cellcolor{best}0.994} & \textbf{\cellcolor{best}48.57} & \textbf{\cellcolor{best}0.992} \\ \hline
Real & 3.02 & 2.17 & 6.77 & 38.32 & 0.974 & 2.133 & 49.80 & 0.993 & 48.25 & 0.990 \\ \hline
\end{tabular}}
\vspace{-4mm}
\end{table}

\cref{tab:downstream_tasks} shows the results. RawGen-based T2R improves performance across all three tasks compared to prior synthetic raw generation approaches. These results indicate that scalable and physically grounded raw synthesis can alleviate data acquisition bottlenecks in device-specific low-level vision tasks.

\paragraph{Generative Image Editing.}
Image editing is more reliable in a scene-referred linear space, where radiometric linearity and a wider dynamic range are preserved, enabling physically grounded operations such as white balance and exposure scaling~\cite{afifi2021cie}. Conventional inverse-ISP methods~\cite{Xing2021invisp, afifi2021cie, reinders2025raw, berdan2025reraw} enable linear-domain editing but are often tuned to specific devices and fixed imaging assumptions, which can lead to failures when inverting heterogeneous sRGB inputs outside the training distribution (\cref{sec:n_to_one}) and degrade editing quality.

Camera-controllable generation frameworks such as Generative Photography~\cite{yuan2025generative} support parameter-driven edits (e.g., white balance, shutter speed, and \textit{f}-number) while preserving scene context, but they operate entirely in the sRGB domain without access to an underlying raw image. Consequently, applying a new parameter setting typically requires rerunning an iterative diffusion process, rather than performing ISP operations on linear sensor data. Moreover, their controllability is limited to the camera parameters explicitly learned by the model, and supporting additional types of edits or new control dimensions generally requires retraining or additional fine-tuning.

In contrast, RawGen decouples content synthesis from rendering by producing a scene-referred linear image from either an input image or a text prompt, and then enabling edits via standard software-ISP operations. After generating the linear representation once, a wide range of edits, including white balance, exposure compensation, and tone mapping, can be applied directly in the linear domain using an arbitrary software ISP pipeline without re-invoking the generative model. This supports efficient exploration of diverse editing operations beyond a fixed set of learned camera parameters.
\section{Conclusion and Discussion}
We present RawGen, a diffusion-based framework for camera-agnostic raw generation that bridges display-referred generative models and scene-referred linear representations. Using a many-to-one objective, RawGen suppresses photo-finishing effects and recovers a canonical XYZ representation consistent across diverse sRGB variants. 
We demonstrated many-to-one linear XYZ reconstruction and device-specific raw synthesis. Moreover, scalable text-to-raw data synthesis enables RawGen to improve downstream low-level vision tasks.

\paragraph{Limitations.} While RawGen generates canonical scene-referred XYZ representations, accurate device-specific raw synthesis also depends on factors beyond color mapping, such as noise characteristics and PSFs. Future work may explore conditioning RawGen on device-specific priors to incorporate more physically grounded sensor modeling, including spatially varying noise, lens shading, and optical blur, further improving device fidelity and physical consistency.
\section*{Acknowledgments}
We thank Ran Zhang for assistance with rendering the generated raw images using neural ISP models. 
\clearpage
\setcounter{page}{16}
\appendix

\begin{center}
    {\Large\bfseries Supplementary Material}\\[1em]
\end{center}
\vspace{1em}

In the main paper, we present RawGen, a method for generating realistic raw images conditioned on text or sRGB inputs. In the supplementary material, we provide additional methodological details and extended experimental results.

\section{Photo-Finishing Simulation Details}
\label{supp:data_generation}

\paragraph{ISP Pipeline.}
Starting from a DNG raw file, we employ a physically grounded ISP simulator based on the software ISP~\cite{Seo2023graphics2raw}. The pipeline processes raw sensor data through the ordered stages:
\texttt{raw} $\rightarrow$ \texttt{normalize} $\rightarrow$ \texttt{lens\_shading\_correction} $\rightarrow$ \texttt{white\_balance} $\rightarrow$ \texttt{demosaic} $\rightarrow$ \texttt{xyz} $\rightarrow$ \texttt{srgb} $\rightarrow$ \texttt{gamma}.
We use edge-aware (EA) demosaicing, adopt a D50 white point for XYZ-to-sRGB conversion, and apply the camera’s embedded forward color matrix (the \texttt{ForwardMatrix} tag in the DNG) for color transformation.

\paragraph{XYZ Anchor.}
The XYZ anchor $I_{\mathrm{XYZ}}$ is obtained by running the ISP pipeline with the DNG’s embedded white-balance coefficients and terminating at the \texttt{xyz} stage, prior to sRGB rendering. No photo-finishing augmentation is applied; the anchor serves as a fixed, illumination-neutral, scene-referred reference for all variants of the same scene.

\paragraph{sRGB Variant Generation.}
For each scene, the ISP pipeline is executed multiple times with independently sampled photo-finishing parameters to produce a set of sRGB renditions $\{I_{\mathrm{sRGB}}^{(n)}\}$. Since parameters are sampled independently per scene, the augmentation space is effectively continuous; the number of renditions reflects computational budget rather than a fixed style count.
Three parameter groups are randomized:

\begin{enumerate}
  \item \textbf{White Balance.}
  Channel gains for red and blue are initialized from the DNG \texttt{AsShotNeutral} metadata and independently perturbed using multiplicative factors $r \sim \mathcal{U}(0.7, 1.3)$ and $b \sim \mathcal{U}(0.7, 1.3)$, whereas the green gain is held constant ($g = 1.0$).

  \item \textbf{Tone Mapping.}
  The ISP output is decoded to linear RGB via the inverse sRGB electro-optical transfer function (OETF). A parametric tone-mapping operator is applied per channel:
  \begin{equation}
      T(E_i) = \frac{(1+\beta)\,E_i^{\,\gamma}}{\beta + E_i^{\,\gamma}},
      \label{eq:lediff}
  \end{equation}
  where $E_i \in [0,1]$ denotes the linear per-channel intensity. Parameters are sampled as $\beta \sim \mathcal{N}(0.6,\,0.1^2)$ (clipped to $[0.1, 2.0]$) and $\gamma \sim \mathcal{N}(0.9,\,0.1^2)$ (clipped to $[0.5, 1.5]$). The tone-mapped result is re-encoded to sRGB using the standard OETF.

  \item \textbf{Contrast.}
  A contrast factor $c \sim \mathcal{U}(0.7, 1.3)$ is applied in the display-referred sRGB domain about a mid-gray pivot (0.5):
  \begin{equation}
      I_{\mathrm{out}} = (I_{\mathrm{in}} - 0.5)\, c + 0.5.
  \end{equation}
\end{enumerate}

\paragraph{Image Resolution and Format.}
Each image is center-cropped to the largest square region and resized to $1024 \times 1024$ using Lanczos interpolation. XYZ anchors are stored as 16-bit PNG to preserve precision, while sRGB variants are stored as 8-bit PNG.

\paragraph{Parameter Ranges.}
\cref{tab:isp_params} summarizes all stochastic parameters.

\begin{table}[!t]
  \caption{Photo-finishing parameter ranges used for sRGB variant generation.}
  \label{tab:isp_params}
  \vspace{-2mm}
  \centering
  \scalebox{0.85}{
  \begin{tabular}{|l|l|l|l|}
  \hline
  Parameter & Distribution & Range & Stage \\ \hline

  Red WB multiplier $r$   
  & Uniform 
  & $[0.7,\ 1.3]$ 
  & White balance \\ \hline

  Blue WB multiplier $b$  
  & Uniform 
  & $[0.7,\ 1.3]$ 
  & White balance \\ \hline

  Tone-map $\beta$        
  & Normal  
  & $\mu{=}0.6,\ \sigma{=}0.1$ (clip $[0.1,\ 2.0]$) 
  & Tone mapping \\ \hline

  Tone-map $\gamma$       
  & Normal  
  & $\mu{=}0.9,\ \sigma{=}0.1$ (clip $[0.5,\ 1.5]$) 
  & Tone mapping \\ \hline

  Contrast $c$            
  & Uniform 
  & $[0.7,\ 1.3]$ 
  & sRGB \\ \hline

  \end{tabular}}
  \vspace{-4mm}
\end{table}

\section{XYZ-to-Camera Raw Mapping}
\label{sec:xyz-to-raw-supp}

To render a decoded CIE XYZ image into a target camera's linear raw-RGB space, we use calibration metadata from a representative DNG of the target camera. 
Specifically, we leverage the DNG-provided matrix pairs under two reference illuminants and interpolate them according to the chosen correlated color temperature (CCT). 
This enables camera-specific rendering while keeping the XYZ generation stage camera-agnostic.

\paragraph{CCT-Based Calibration Matrix Interpolation.}
Let $\mathbf{C}_1,\mathbf{C}_2$ denote the \texttt{Color\allowbreak Matrix} under two calibration illuminants, and $\mathbf{F}_1,\mathbf{F}_2$ denote the corresponding \texttt{ForwardMatrix} extracted from the DNG file. 
Given a target CCT $T$, we compute an interpolation weight using the reciprocal-temperature rule:
\begin{equation}
g = \mathrm{clip}\!\left(\frac{1/T - 1/T_1}{1/T_2 - 1/T_1},\, 0,\, 1\right),
\end{equation}
and form interpolated matrices
\begin{equation}
\mathbf{C}(T) = g\,\mathbf{C}_1 + (1-g)\,\mathbf{C}_2,\quad
\mathbf{F}(T) = g\,\mathbf{F}_1 + (1-g)\,\mathbf{F}_2.
\end{equation}

\paragraph{XYZ to Camera Raw-RGB.}
Given a decoded linear XYZ image $\widehat{I}_{\mathrm{XYZ}}$, we first map it to a white-balanced camera RGB by inverting the forward model:
\begin{equation}
I_{\mathrm{WB}} = \mathrm{clip}\!\left(\widehat{I}_{\mathrm{XYZ}} \cdot \mathbf{F}(T)^{-T},\, 0,\, 1\right),
\end{equation}
where the matrix is applied per pixel. 
We then convert an illuminant in XYZ, $\boldsymbol{\ell}_{\mathrm{XYZ}}$, into camera RGB via the color matrix,
\begin{equation}
\boldsymbol{\ell}_{\mathrm{RGB}} = \mathbf{C}(T)\,\boldsymbol{\ell}_{\mathrm{XYZ}},
\end{equation}
normalize it by the green channel to obtain a relative gain vector, and apply it to obtain an illuminated camera RGB, which we treat as the target camera's linear raw-RGB output:
\begin{equation}
I_{\mathrm{raw}} = \mathrm{clip}\!\left(I_{\mathrm{WB}} \odot \boldsymbol{\ell}_{\mathrm{RGB}},\, 0,\, 1\right).
\end{equation}
Unless otherwise noted, we export $I_{\mathrm{raw}}$ as a normalized 16-bit image for dataset generation.

\paragraph{Heteroscedastic Noise Simulation.}
\label{sec:noise-supp}
\begin{figure*}[!t]
  \centering
  \includegraphics[width=\columnwidth]{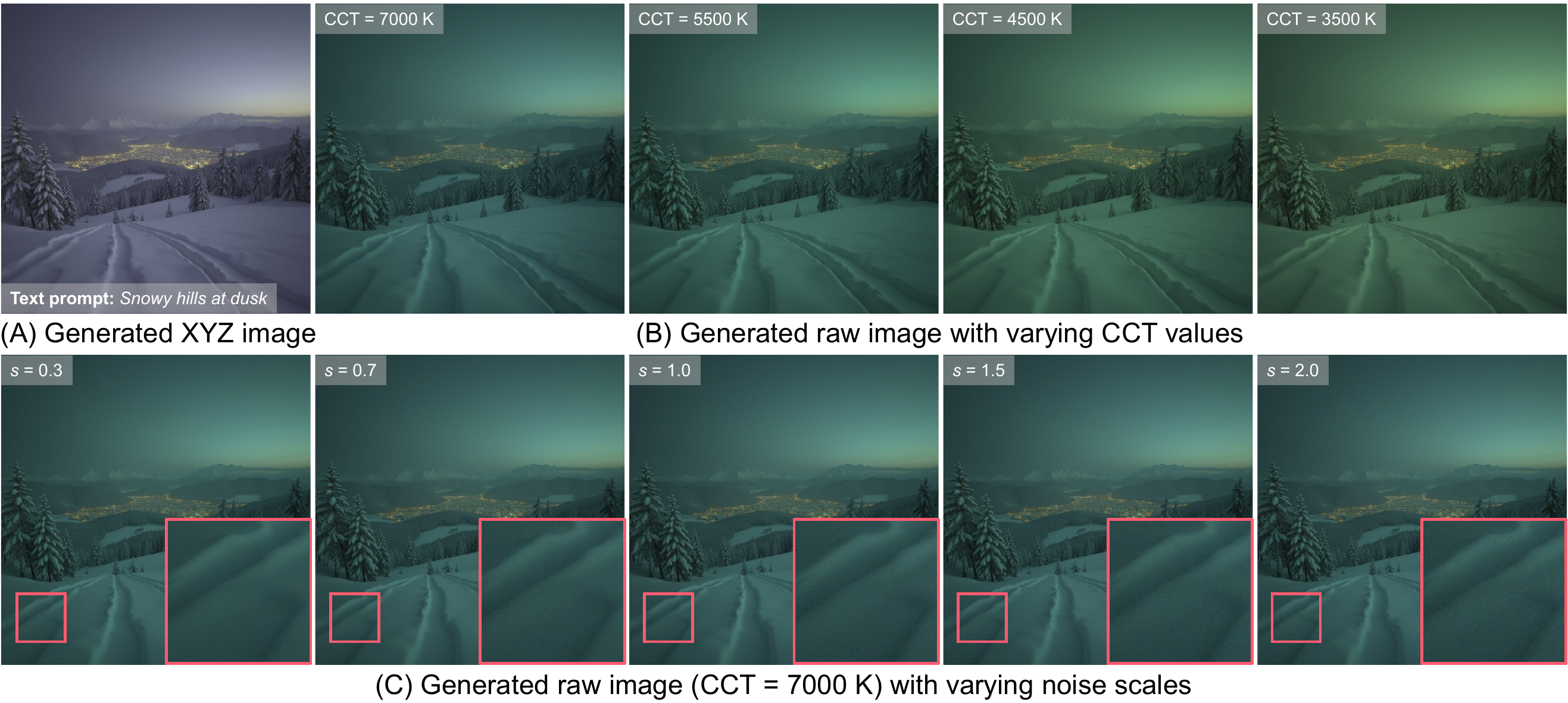}
\vspace{-4mm}
\caption{Variations of raw images generated from the CIE XYZ image shown in (A). 
(B) shows raw images mapped to the Samsung Galaxy S24 main camera raw space under different correlated color temperatures (CCTs). 
(C) shows raw images mapped with a fixed CCT of 7000\,K and different noise strength factor values $s$. All images are gamma-corrected for visualization throughout this supplementary material.}

  \label{fig:wb-noise}
\end{figure*}
We optionally simulate target sensor noise by applying a heteroscedastic noise model to the resulting raw image $I_{\mathrm{raw}}$.
For each color channel, the noise variance is modeled as a linear function of the signal intensity, scaled by a global noise strength factor $s$:
\begin{equation}
\sigma_c^2(I_{\mathrm{raw}}) = s \left( \alpha_c \, I_{\mathrm{raw}}^{(c)} + \beta_c \right),
\end{equation}
where $\alpha_c$ and $\beta_c$ represent the signal-dependent (shot) and signal-independent (read) noise components for channel $c$, respectively, and $s$ controls the overall noise magnitude.
The parameters $\alpha_c$ and $\beta_c$ are obtained from camera noise calibration \cite{foi2008practical} or derived from the camera's noise profile \cite{abdelhamed2018high}.

Zero-mean Gaussian noise is then sampled according to this variance and added to the input image, producing the noisy output:
\begin{equation}
I_{\mathrm{raw}}' = I_{\mathrm{raw}} + \mathcal{N}\!\left(0,\, \sigma^2(I_{\mathrm{raw}})\right).
\end{equation}

\cref{fig:wb-noise} shows an example of mapping a generated CIE XYZ image to the raw space of the Samsung Galaxy S24 main camera under different CCTs and noise strength factor values.

\section{Downstream Task Setup Details}
\label{sec:supp_downstream_setup}

\paragraph{Illumination Estimation.}
We follow Graphics2RAW and evaluate illuminant estimation on the nine-camera NUS dataset~\cite{cheng2014illuminant}. We generate 3K synthetic raw images from text prompts to increase scene diversity. The network architecture~\cite{Gong2019ConvolutionalMA} and optimization settings are kept identical to the reference protocol, and performance is measured using angular error. This controlled setup isolates the impact of RawGen's scalable generative data on learning color constancy.

\paragraph{Neural ISP.}
To assess RawGen for neural ISP training, we use the nighttime dataset~\cite{Punnappurath_2022_CVPR} and testing configuration introduced in Graphics2RAW. 3K synthetic raw images generated with RawGen are paired with rendered sRGB targets and used for supervision. The ISP architecture~\cite{ronneberger2015u}, loss functions, and evaluation metrics are kept fixed, and testing is performed on real raw inputs. This setup evaluates whether RawGen provides effective supervision for real-world ISP rendering.

\paragraph{Image Denoising.}
We further evaluate RawGen on raw-domain denoising using the nighttime dataset~\cite{Punnappurath_2022_CVPR} and protocol introduced in Graphics2RAW. For training, we synthesize 3K clean linear images from text prompts and generate noisy counterparts with the same heteroscedastic Gaussian noise model as the baseline. The denoiser architecture~\cite{zamir2022restormer} and training configuration remain unchanged, and testing is conducted exclusively on real noisy captures at ISOs 1600 and 3200. This controlled setup isolates the benefit of RawGen for improving noise-robust learning through scalable training data.

\section{Additional Results}
\label{sec:results-supp}

\begin{figure}[!t]
  \centering
  \includegraphics[width=\columnwidth]{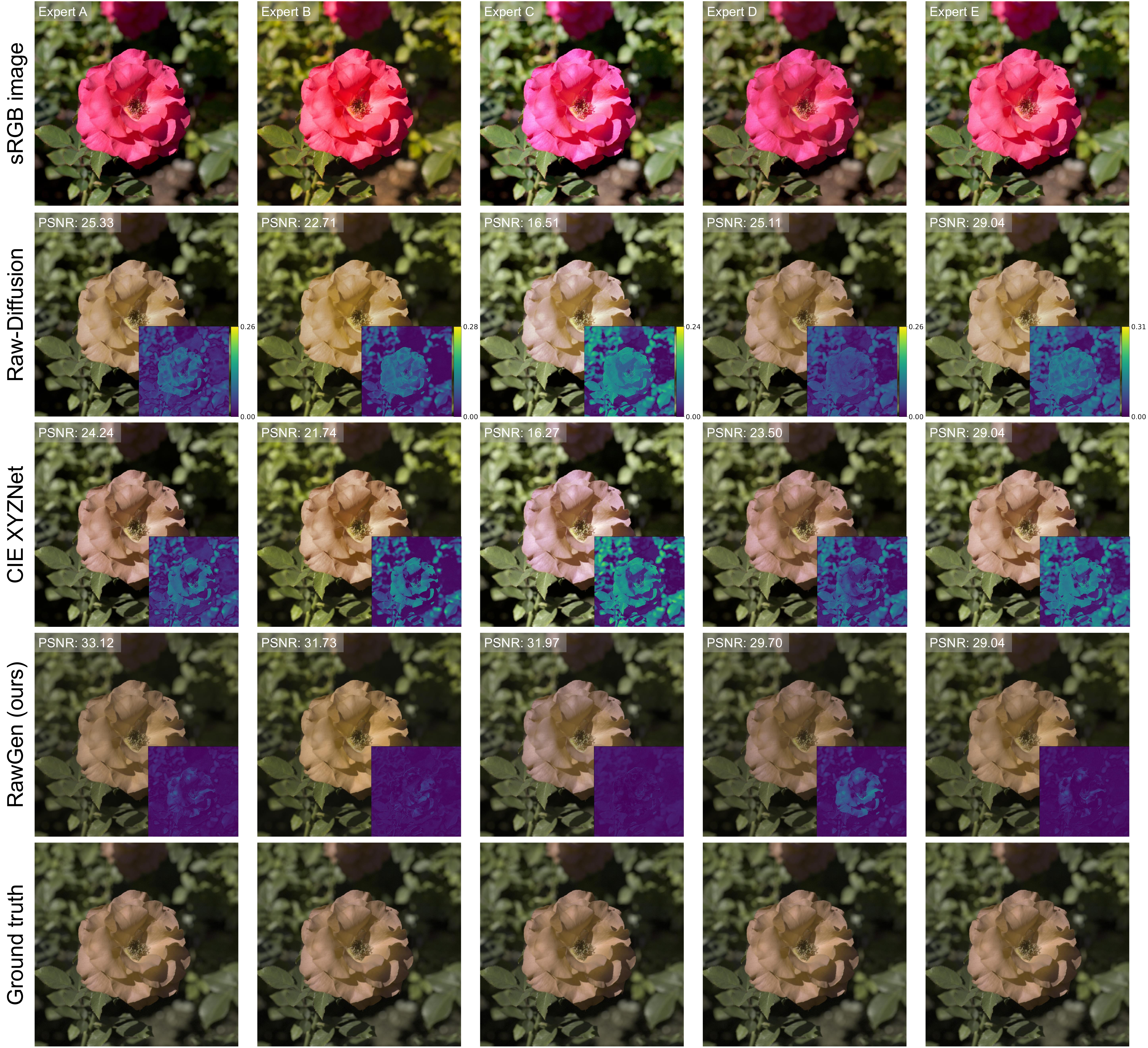}
\vspace{-4mm}
\caption{Additional CIE XYZ reconstruction results.
Shown is an sRGB input image rendered with different rendering preferences (Expert A--E) and the corresponding CIE XYZ reconstructions produced by Raw-Diffusion \cite{reinders2025raw}, CIE XYZ Net \cite{afifi2021cie}, and our RawGen. The last row shows the ground-truth CIE XYZ image.}
  \label{fig:xyz-qualitative-supp-1}
\end{figure}

\begin{figure}[!t]
  \centering
  \includegraphics[width=\columnwidth]{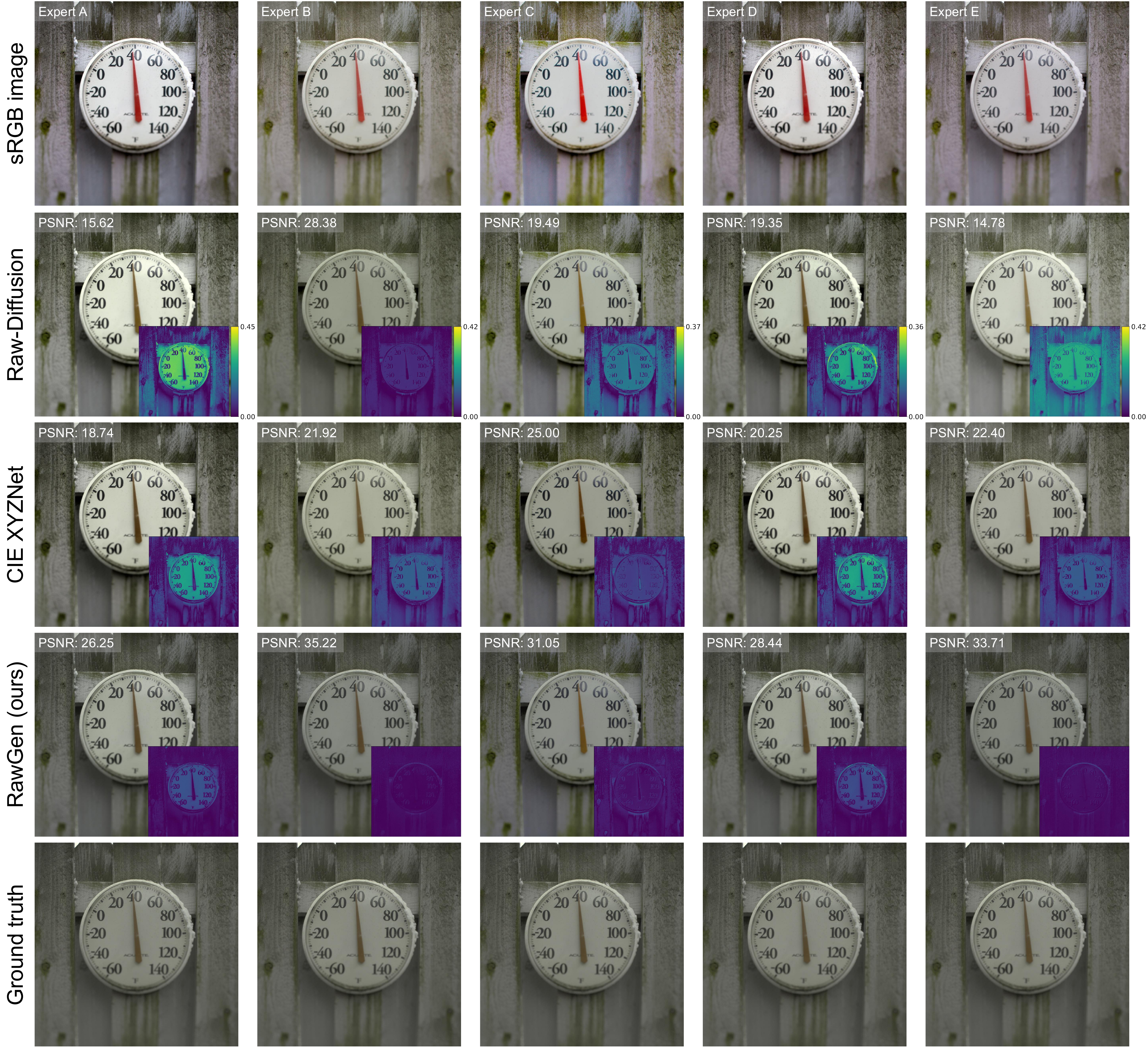}
\vspace{-4mm}
\caption{Additional CIE XYZ reconstruction results.
Shown is an sRGB input image rendered with different rendering preferences (Expert A--E) and the corresponding CIE XYZ reconstructions produced by Raw-Diffusion \cite{reinders2025raw}, CIE XYZ Net \cite{afifi2021cie}, and our RawGen. The last row shows the ground-truth CIE XYZ image.}
  \label{fig:xyz-qualitative-supp-3}
\end{figure}

\begin{figure}[!t]
  \centering
  \includegraphics[width=\columnwidth]{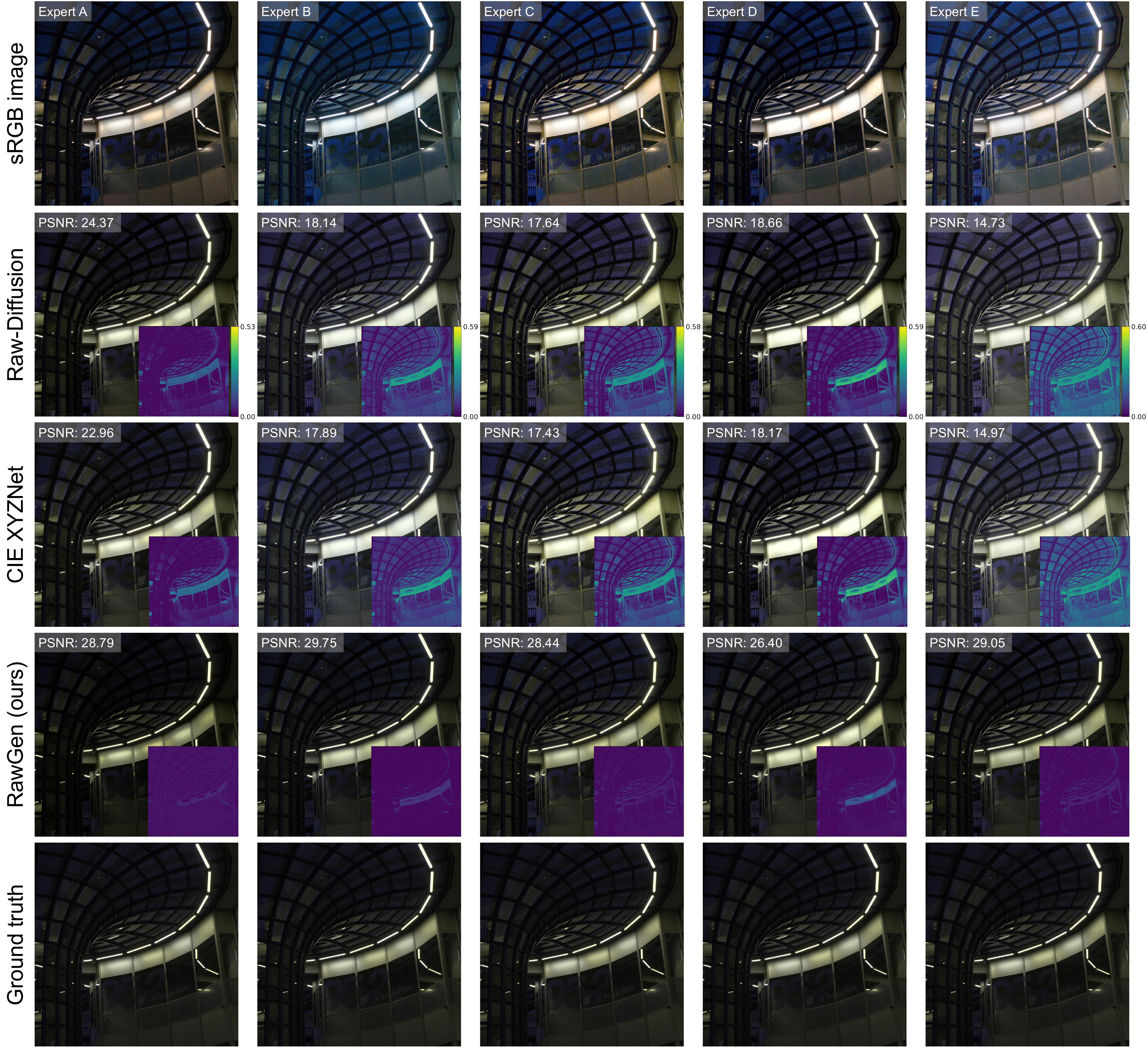}
\vspace{-4mm}
\caption{Additional CIE XYZ reconstruction results.
Shown is an sRGB input image rendered with different rendering preferences (Expert A--E) and the corresponding CIE XYZ reconstructions produced by Raw-Diffusion \cite{reinders2025raw}, CIE XYZ Net \cite{afifi2021cie}, and our RawGen. The last row shows the ground-truth CIE XYZ image.}
  \label{fig:xyz-qualitative-supp-4}
\end{figure}

\begin{figure}[!t]
  \centering
  \includegraphics[width=\columnwidth]{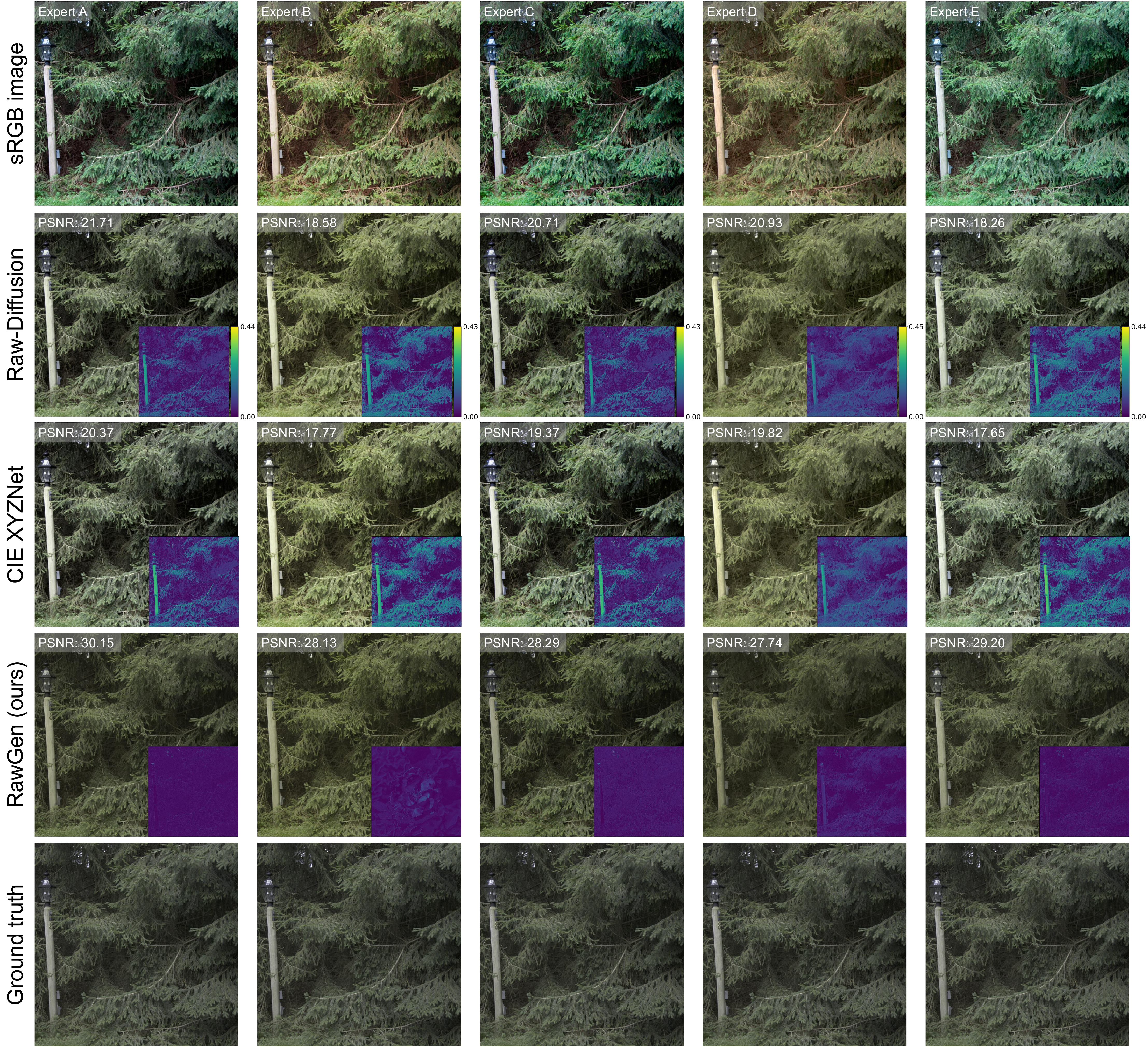}
\vspace{-4mm}
\caption{
Additional CIE XYZ reconstruction results.
Shown is an sRGB input image rendered with different rendering preferences (Expert A--E) and the corresponding CIE XYZ reconstructions produced by Raw-Diffusion \cite{reinders2025raw}, CIE XYZ Net \cite{afifi2021cie}, and our RawGen. The last row shows the ground-truth CIE XYZ image.
}

  \label{fig:xyz-qualitative-supp-5}
\end{figure}

\begin{figure}[!t]
  \centering
  \includegraphics[width=\columnwidth]{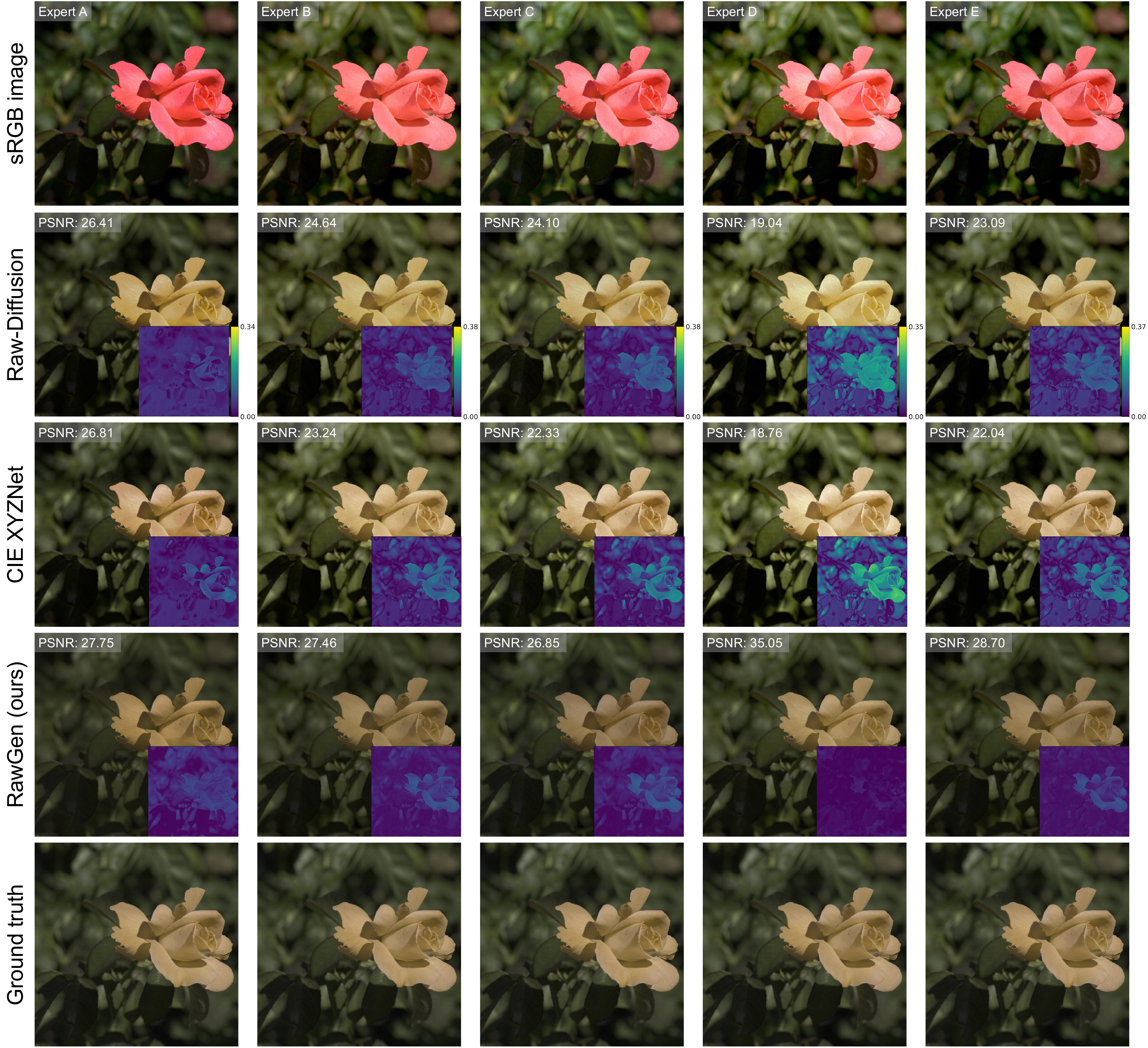}
\vspace{-4mm}
\caption{Additional CIE XYZ reconstruction results.
Shown is an sRGB input image rendered with different rendering preferences (Expert A--E) and the corresponding CIE XYZ reconstructions produced by Raw-Diffusion \cite{reinders2025raw}, CIE XYZ Net \cite{afifi2021cie}, and our RawGen. The last row shows the ground-truth CIE XYZ image.}
  \label{fig:xyz-qualitative-supp-6}
\end{figure}

\paragraph{CIE XYZ Reconstruction.} In Sec.~4.2 of the main paper, we report quantitative results on CIE XYZ reconstruction under unseen photo-finishing styles using the MIT-Adobe FiveK dataset~\cite{fivek}, where each scene is edited by five experts (A–E) with distinct aesthetic preferences. Here, we provide additional qualitative comparisons in \cref{fig:xyz-qualitative-supp-1,fig:xyz-qualitative-supp-3,fig:xyz-qualitative-supp-4,fig:xyz-qualitative-supp-5,fig:xyz-qualitative-supp-6}.

\begin{figure}[!t]
  \centering
  \includegraphics[width=\columnwidth]{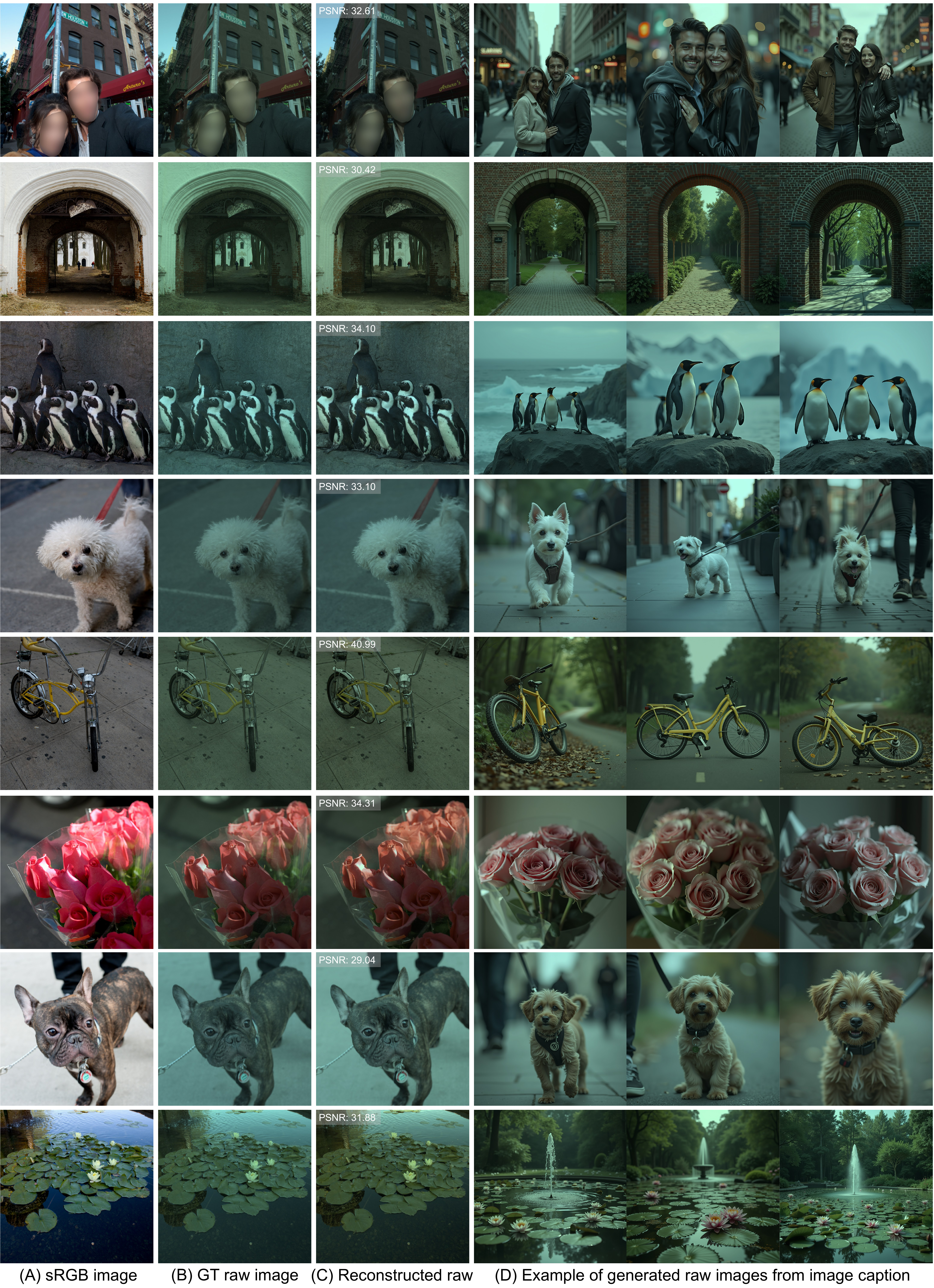}
\vspace{-4mm}
\caption{
Additional raw image reconstruction and generation results.
Shown are the input sRGB images (A) and corresponding ground-truth raw images (B). Our reconstructed raw results are shown in (C). Additional generated raw images of the same semantic scene are shown in (D), where InstructBLIP-2 \cite{dai2023instructblip} is used to generate text descriptions from image (A).
}
  \label{fig:qual-results-supp}
\end{figure}

\begin{figure*}[!t]
  \centering
  \includegraphics[width=\columnwidth]{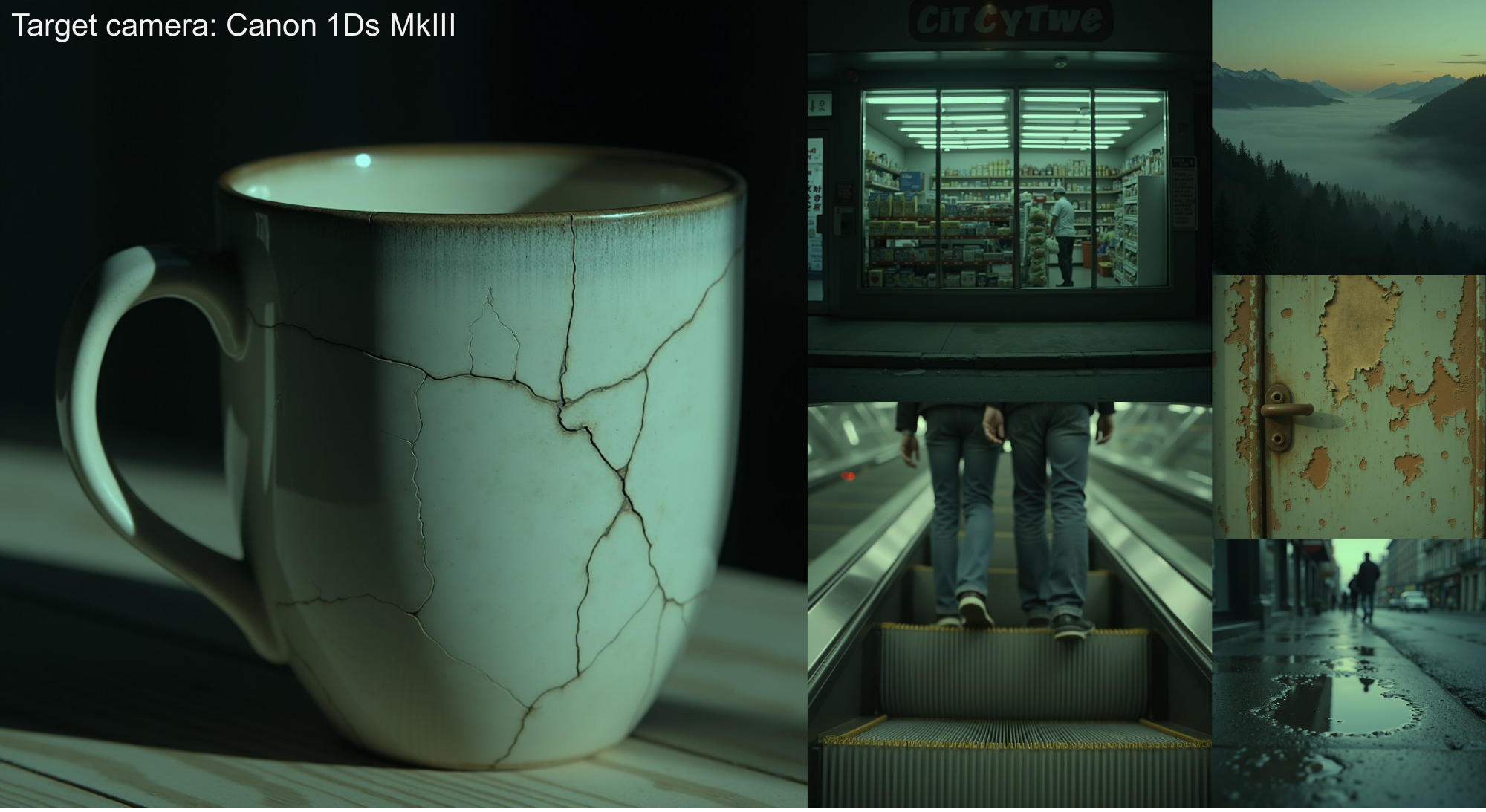}
  \vspace{-2mm}
\caption{Examples of generated raw images for the Canon 1Ds MkIII DSLR camera. The images were generated using the following text prompts: 1) \texttt{Cracked ceramic mug with visible hairline fractures, natural lighting}, 2) \texttt{convenience store at 2am, viewed through a glass facade}, 3) \texttt{commuters stepping onto an escalator at the exact moment of boarding}, 4) \texttt{mountain valley just before sunrise with rising ground fog}, 5) \texttt{peeling paint on a weathered metal door}, and 6) \texttt{city sidewalk just as a shallow puddle finishes forming}.
}
  \label{fig:raw-examples-1}
\end{figure*}

\begin{figure*}[!t]
  \centering
  \includegraphics[width=\columnwidth]{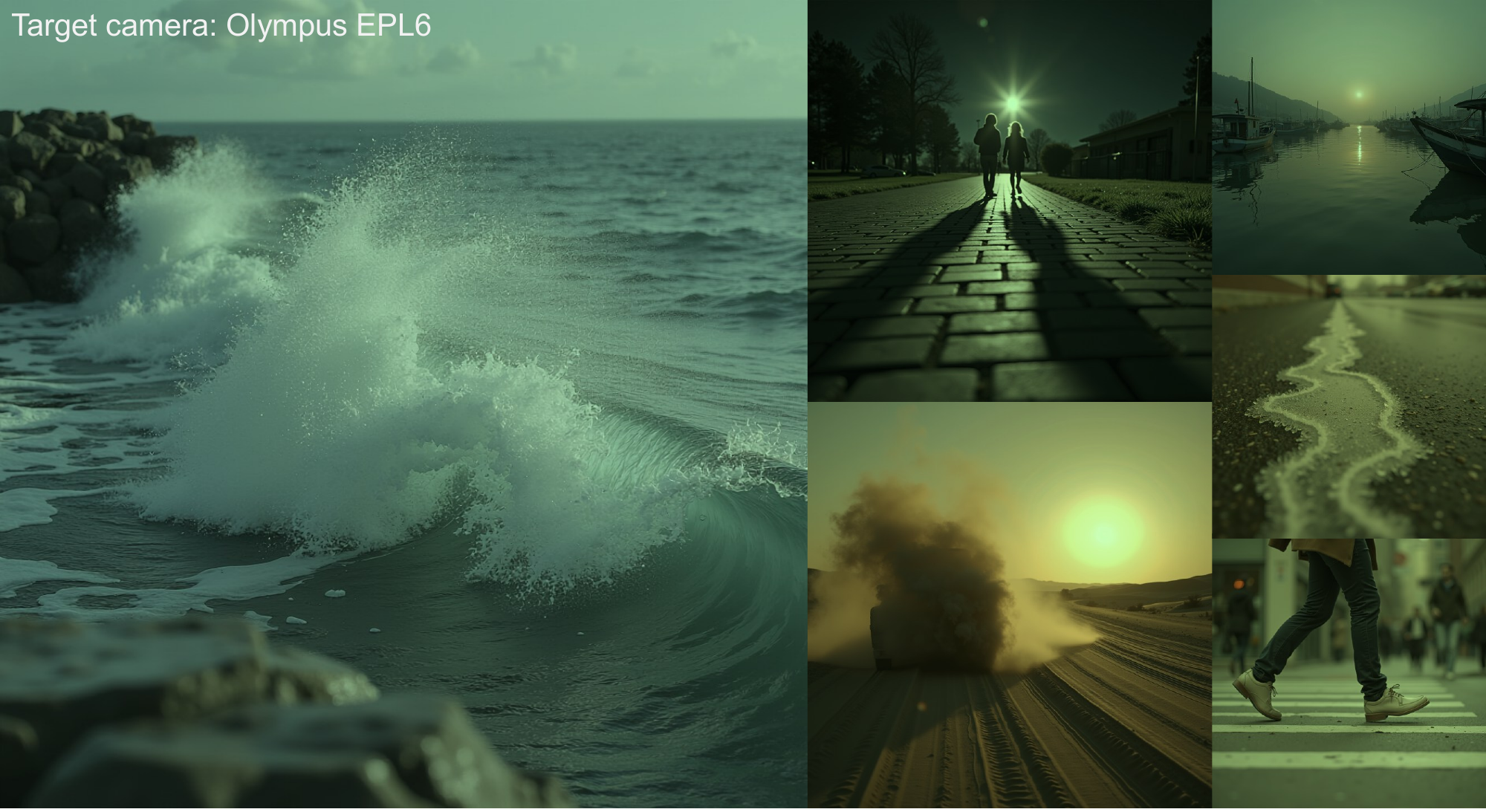}
  \vspace{-2mm}
\caption{Examples of generated raw images for the Olympus EPL6 DSLR camera. The images were generated using the following text prompts: 1) \texttt{ocean waves striking a seawall at the exact moment of impact}, 2) \texttt{shadow band phenomenon in the final seconds before totality}, 3) \texttt{wind-driven sand crossing a desert road at sunset}, 4) \texttt{coastal fishing village harbor at dawn before activity begins}, 5) \texttt{suburban parking lot after freezing rain}, and 6) \texttt{pedestrian mid-step with both feet briefly off the ground}.
}
  \label{fig:raw-examples-2}
\end{figure*}

\begin{figure*}[!t]
  \centering
  \includegraphics[width=\columnwidth]{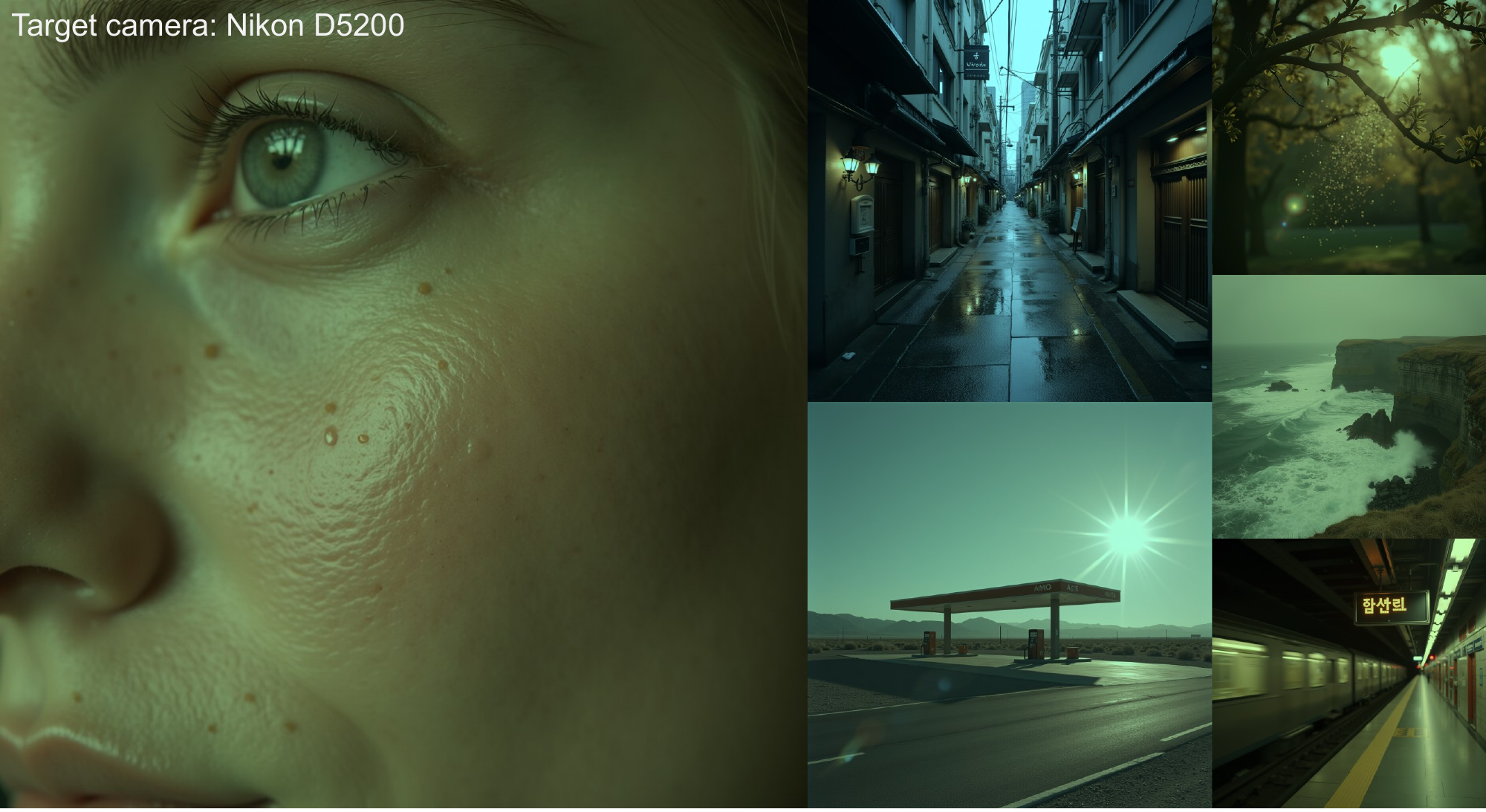}
  \vspace{-2mm}
\caption{Examples of generated raw images for the Nikon D5200 DSLR camera. The images were generated using the following text prompts: 1) \texttt{Close-up of human skin with visible pores and fine hairs}, 2) \texttt{narrow alley after rainfall}, 3) \texttt{roadside gas station in midday heat shimmer}, 4) \texttt{tree pollen drifting in golden backlight}, 5) \texttt{coastal cliff facing the ocean in strong onshore winds}, and 6) \texttt{subway platform during non-peak hours}.
}
  \label{fig:raw-examples-3}
\end{figure*}

\begin{figure*}[!t]
  \centering
  \includegraphics[width=\columnwidth]{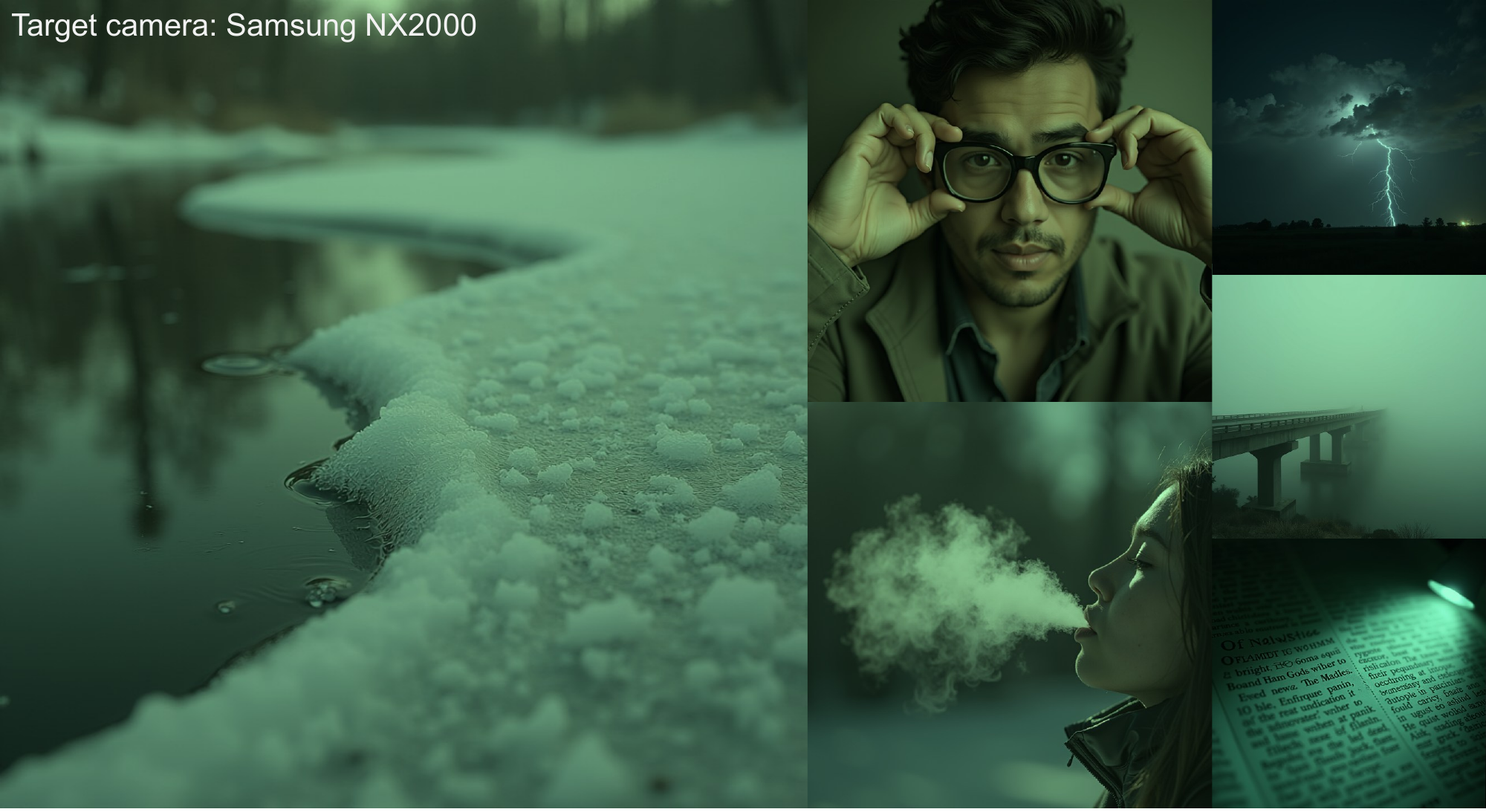}
  \vspace{-2mm}
\caption{Examples of generated raw images for the Samsung NX2000 DSLR camera. The images were generated using the following text prompts: 1) \texttt{Thin ice forming across a pond at the transition point}, 2) \texttt{person adjusting glasses at the precise midpoint of the motion}, 3) \texttt{visible breath at the exact moment of exhale in cold air}, 4) \texttt{lightning illuminating low cloud cover without direct bolt visibility}, 5) \texttt{coastal fog rolling over a highway overpass}, and 6) \texttt{newspaper print under fluorescent lighting with slight glare}.
}
  \label{fig:raw-examples-4}
\end{figure*}

\paragraph{sRGB-to-Raw Synthesis.} In Sec.~4.3 of the main paper, we compared RawGen-synthesized raw images with ground-truth raw data and presented additional raw samples for the same semantic scene generated from image descriptions extracted from the input sRGB images using InstructBLIP-2~\cite{dai2023instructblip}. Additional results are shown in \cref{fig:qual-results-supp}. We further present Text-to-Raw (T2R)-based raw generation results, where the CIE XYZ images generated by RawGen are mapped to arbitrary target camera raw domains (\cref{fig:raw-examples-1,fig:raw-examples-2,fig:raw-examples-3,fig:raw-examples-4}).

\paragraph{Compatibility on Pre-Trained ISP.} In Sec.~4.3 of the main paper, we evaluate the compatibility of RawGen-generated raw images with neural ISP models trained exclusively on real raw data from the Samsung Galaxy S24 camera (S24 dataset~\cite{afifi2025time}).

\begin{figure}[!t]
  \centering
  \includegraphics[width=\columnwidth]{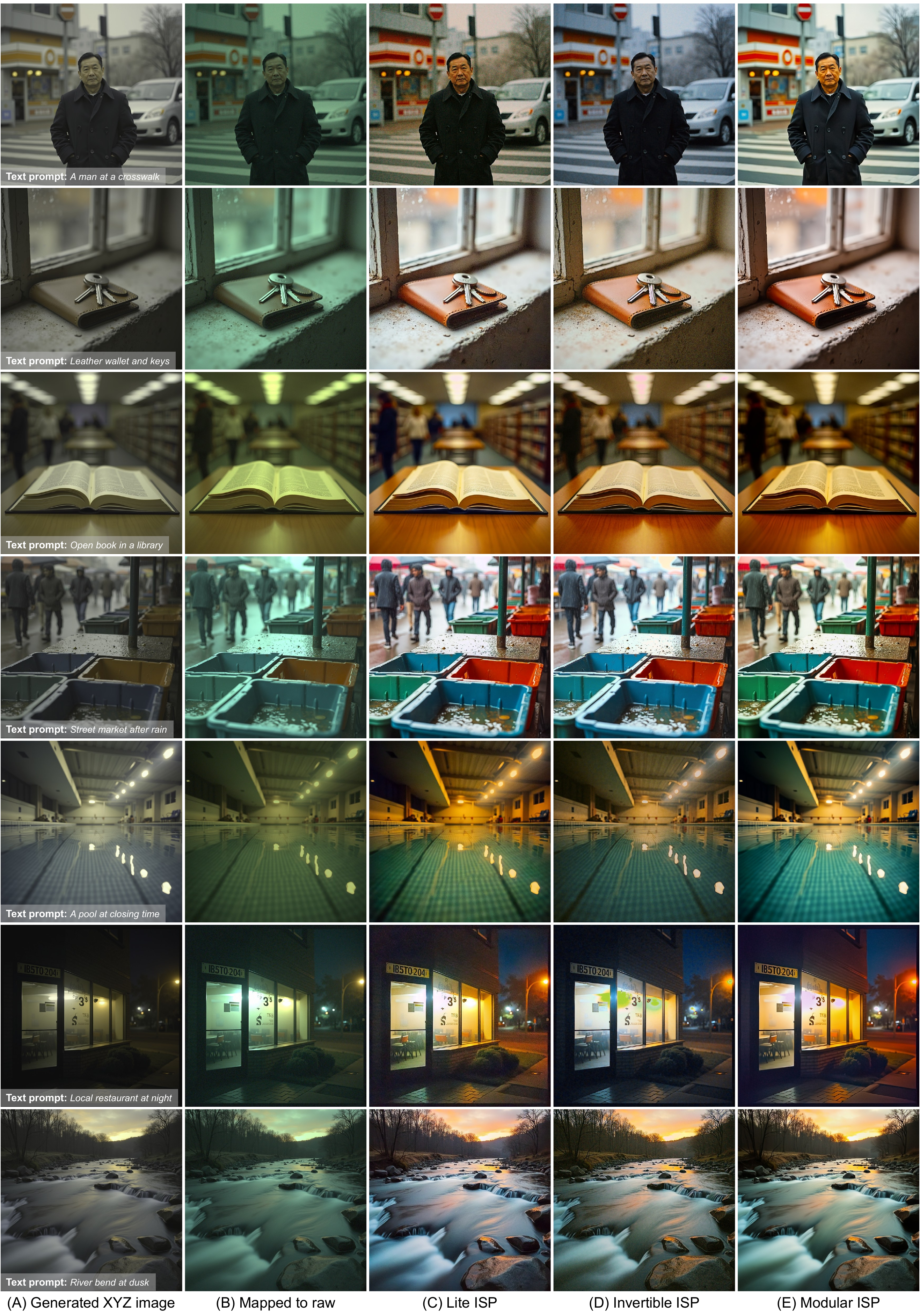}
\vspace{-4mm}
\caption{Rendering results of pre-trained neural ISPs on real data from the Samsung S24 main camera (S24 dataset~\cite{afifi2025time}). (A) CIE XYZ images generated by our RawGen. (B) Images mapped to the S24 main camera raw space with synthetic noise. (C) Result of Lite ISP~\cite{zhang2021learning}. (D) Result of Invertible  ISP~\cite{Xing2021invisp}. (E) Result of Modular Neural ISP~\cite{afifi2025modular}.}
  \label{fig:pre-trained-neural-isp-supp}
\end{figure}

Here, we provide additional compatibility results for Lite ISP~\cite{zhang2021learning}, Invertible ISP~\cite{Xing2021invisp}, and Modular Neural ISP~\cite{afifi2025modular}. All ISP models are fed RawGen-generated raw images with synthesized sensor noise, and their rendered sRGB outputs are shown in \cref{fig:pre-trained-neural-isp-supp}. The visually plausible renderings—despite known generalization limitations when raw distributions deviate from the training camera~\cite{afifi2021semi, perevozchikov2024rawformer}—indicate close alignment between RawGen-generated raw images and real S24 training data.

\begin{figure*}[!t]
  \centering
  \includegraphics[width=0.95\columnwidth]{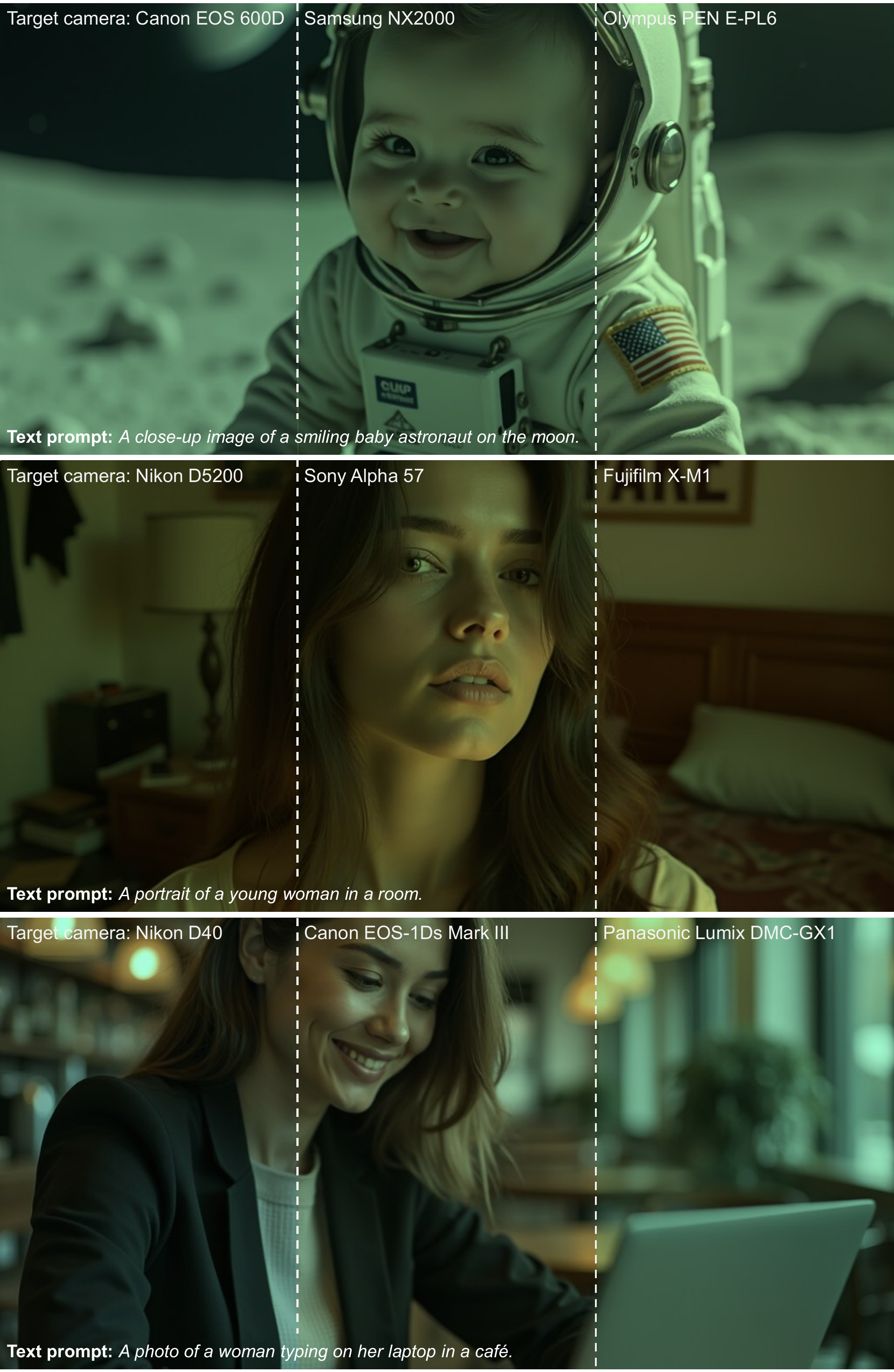}
  \vspace{-2mm}
\caption{Additional results of our RawGen, generated from the text prompts shown at the bottom of each image. For each prompt, we show generated raw images for three target cameras.}
  \label{fig:raw-examples-0}
\end{figure*}

\clearpage  

%
%
\bibliographystyle{splncs04}
\bibliography{main}

@String(CVPR= {IEEE Conf. Comput. Vis. Pattern Recog.})

@String(ICCV= {Int. Conf. Comput. Vis.})

@String(ECCV= {Eur. Conf. Comput. Vis.})

@String(BMVC= {Brit. Mach. Vis. Conf.})

@String(TOG= {ACM Trans. Graph.})

@String(ICLR = {Int. Conf. Learn. Represent.})

@String(AAAI = {AAAI})

@String(CVPRW= {IEEE Conf. Comput. Vis. Pattern Recog. Worksh.})

@String(CVPR  = {CVPR})

@String(ICCV  = {ICCV})

@String(ECCV  = {ECCV})

@String(BMVC  =	{BMVC})

@String(TOG   = {ACM TOG})

@String(ICLR  = {ICLR})

@String(CVPRW= {CVPRW})

@inproceedings{saharia2022photorealistic,
title     = {Photorealistic Text-to-Image Diffusion Models with Deep Language Understanding},
author    = {Saharia, Chitwan and Chan, William and Saxena, Saurabh and Li, Lala and Whang, Jay and others},
booktitle = {NeurIPS},
year      = {2022}
}

@inproceedings{rombach2022high,
title     = {High-Resolution Image Synthesis with Latent Diffusion Models},
author    = {Rombach, Robin and Blattmann, Andreas and Lorenz, Dominik and Esser, Patrick and Ommer, Bjorn},
booktitle = {CVPR},
year      = {2022}
}

@article{ramesh2022hierarchical,
title   = {Hierarchical Text-Conditional Image Generation with {CLIP} Latents},
author  = {Ramesh, Aditya and Dhariwal, Prafulla and Nichol, Alex and Chu, Casey and Chen, Mark},
journal = {arXiv preprint arXiv:2204.06125},
year    = {2022}
}

@article{podell2023sdxl,
title   = {{SDXL}: Improving Latent Diffusion Models for High-Resolution Image Synthesis},
author  = {Podell, David and Weng, Chung-I and Onoe, Yasufumi and others},
journal = {arXiv preprint arXiv:2307.01952},
year    = {2023}
}

@inproceedings{zamir2020cycleisp,
title     = {{CycleISP}: Real Image Restoration via Improved Data Synthesis},
author    = {Zamir, Syed Waqas and Arora, Aditya and Khan, Salman and others},
booktitle = {CVPR},
year      = {2020}
}

@inproceedings{brooks2019unprocessing,
title     = {Unprocessing Images for Learned Raw Denoising},
author    = {Brooks, Tim and Mildenhall, Ben and Xue, Tianfan and Chen, Jiawen and Sharlet, Dillon and Barron, Jonathan T.},
booktitle = {CVPR},
year      = {2019}
}

@article{jiang2021enlightengan,
  title={{EnlightenGAN}: Deep light enhancement without paired supervision},
  author={Jiang, Yifan and Gong, Xinyu and Liu, Ding and Cheng, Yu and Fang, Chen and Shen, Xiaohui and Yang, Jianchao and Zhou, Pan and Wang, Zhangyang},
  journal={IEEE Transactions on Image Processing},
  volume={30},
  pages={2340--2349},
  year={2021},
}

@InProceedings{Seo2023graphics2raw,
    author    = {Seo, Donghwan and Punnappurath, Abhijith and Zhao, Luxi and Abdelhamed, Abdelrahman and Tedla, Sai Kiran and Park, Sanguk and Choe, Jihwan and Brown, Michael S.},
    title     = {{Graphics2RAW}: Mapping Computer Graphics Images to Sensor RAW Images},
    booktitle = {ICCV},
    year      = {2023},
}

@article{afifi2021cie,
  title={{CIE} {XYZ} {N}et: Unprocessing images for low-level computer vision tasks},
  author={Afifi, Mahmoud and Abdelhamed, Abdelrahman and Abuolaim, Abdullah and Punnappurath, Abhijith and Brown, Michael S},
  journal={IEEE Transactions on Pattern Analysis and Machine Intelligence},
  volume={44},
  number={9},
  pages={4688--4700},
  year={2021},
}

@InProceedings{Xing2021invisp,
    author    = {Xing, Yazhou and Qian, Zian and Chen, Qifeng},
    title     = {Invertible Image Signal Processing},
    booktitle = {CVPR},
    year      = {2021},
}

@inproceedings{berdan2025reraw,
  title={{ReRAW}: {RGB}-to-raw image reconstruction via stratified sampling for efficient object detection on the edge},
  author={Berdan, Radu and Besbinar, Beril and Reinders, Christoph and Otsuka, Junji and Iso, Daisuke},
  booktitle={CVPR},
  year={2025}
}

@article{zhang2023adding,
title   = {Adding Conditional Control to Text-to-Image Diffusion Models},
author  = {Zhang, Lvmin and Rao, Anyi and Agrawala, Maneesh},
journal = {arXiv preprint arXiv:2302.05543},
year    = {2023}
}

@inproceedings{dang2015raise,
  title={Raise: A raw images dataset for digital image forensics},
  author={Dang-Nguyen, Duc-Tien and Pasquini, Cecilia and Conotter, Valentina and Boato, Giulia},
  booktitle={ACM Multimedia Systems},
  year={2015}
}

@article{cheng2014illuminant,
  title={Illuminant estimation for color constancy: {W}hy spatial-domain methods work and the role of the color distribution},
  author={Cheng, Dongliang and Prasad, Dilip K and Brown, Michael S},
  journal={Journal of the Optical Society of America A},
  volume={31},
  number={5},
  pages={1049--1058},
  year={2014},
}

@article{song2019generative,
  title={Generative modeling by estimating gradients of the data distribution},
  author={Song, Yang and Ermon, Stefano},
  journal={NeurIPS},
  year={2019}
}

@article{ho2020denoising,
  title={Denoising diffusion probabilistic models},
  author={Ho, Jonathan and Jain, Ajay and Abbeel, Pieter},
  journal={NeurIPS},
  year={2020}
}

@article{song2020score,
  title={Score-based generative modeling through stochastic differential equations},
  author={Song, Yang and Sohl-Dickstein, Jascha and Kingma, Diederik P and Kumar, Abhishek and Ermon, Stefano and Poole, Ben},
  journal={arXiv preprint arXiv:2011.13456},
  year={2020}
}

@inproceedings{nichol2021improved,
  title={Improved denoising diffusion probabilistic models},
  author={Nichol, Alexander Quinn and Dhariwal, Prafulla},
  booktitle={ICML},
  year={2021},
}

@article{song2020denoising,
  title={Denoising diffusion implicit models},
  author={Song, Jiaming and Meng, Chenlin and Ermon, Stefano},
  journal={arXiv preprint arXiv:2010.02502},
  year={2020}
}

@article{dhariwal2021diffusion,
  title={Diffusion models beat {GANs} on image synthesis},
  author={Dhariwal, Prafulla and Nichol, Alexander},
  journal={NeurIPS},
  year={2021}
}

@article{karras2022elucidating,
  title={Elucidating the design space of diffusion-based generative models},
  author={Karras, Tero and Aittala, Miika and Aila, Timo and Laine, Samuli},
  journal={NeurIPS},
  year={2022}
}

@article{ho2022classifier,
  title={Classifier-free diffusion guidance},
  author={Ho, Jonathan and Salimans, Tim},
  journal={arXiv preprint arXiv:2207.12598},
  year={2022}
}

@article{song2023consistency,
  title={Consistency Models},
  author={Song, Yang and Dhariwal, Prafulla and Chen, Mark and Sutskever, Ilya},
  journal={arXiv e-prints},
  year={2023}
}

@article{pernias2023wurstchen,
  title={W{\"u}rstchen: An efficient architecture for large-scale text-to-image diffusion models},
  author={Pernias, Pablo and Rampas, Dominic and Richter, Mats L and Pal, Christopher J and Aubreville, Marc},
  journal={arXiv preprint arXiv:2306.00637},
  year={2023}
}

@inproceedings{ramesh2021zero,
  title={Zero-shot text-to-image generation},
  author={Ramesh, Aditya and Pavlov, Mikhail and Goh, Gabriel and Gray, Scott and Voss, Chelsea and Radford, Alec and Chen, Mark and Sutskever, Ilya},
  booktitle={ICML},
  year={2021},
}

@techreport{betker2023improving,
  title        = {Improving Image Generation with Better Captions},
  author       = {Betker, James and Goh, Gabriel and Jing, Li and Brooks, Tim and Wang, Jianfeng and Li, Linjie and Ouyang, Long and Zhuang, Juntang and Lee, Joyce and Guo, Yufei and Manassra, Wesam and Dhariwal, Prafulla and Chu, Casey and Jiao, Yunxin and Ramesh, Aditya},
  institution  = {OpenAI},
  year         = {2023},
  url          = {https://cdn.openai.com/papers/dall-e-3.pdf},
  note         = {Technical report},
}

@article{zhou2024transfusion,
  title={Transfusion: Predict the next token and diffuse images with one multi-modal model},
  author={Zhou, Chunting and Yu, Lili and Babu, Arun and Tirumala, Kushal and Yasunaga, Michihiro and Shamis, Leonid and Kahn, Jacob and Ma, Xuezhe and Zettlemoyer, Luke and Levy, Omer},
  journal={arXiv preprint arXiv:2408.11039},
  year={2024}
}

@article{dai2023emu,
  title={Emu: Enhancing image generation models using photogenic needles in a haystack},
  author={Dai, Xiaoliang and Hou, Ji and Ma, Chih-Yao and Tsai, Sam and Wang, Jialiang and Wang, Rui and Zhang, Peizhao and Vandenhende, Simon and Wang, Xiaofang and Dubey, Abhimanyu and others},
  journal={arXiv preprint arXiv:2309.15807},
  year={2023}
}

@inproceedings{esser2024scaling,
  title={Scaling rectified flow transformers for high-resolution image synthesis},
  author={Esser, Patrick and Kulal, Sumith and Blattmann, Andreas and Entezari, Rahim and M{\"u}ller, Jonas and Saini, Harry and Levi, Yam and Lorenz, Dominik and Sauer, Axel and Boesel, Frederic and others},
  booktitle={ICML},
  year={2024}
}

@article{batifol2025flux,
  title={{FLUX. 1 Kontext}: Flow Matching for In-Context Image Generation and Editing in Latent Space},
  author={Batifol, Stephen and Blattmann, Andreas and Boesel, Frederic and Consul, Saksham and Diagne, Cyril and Dockhorn, Tim and English, Jack and English, Zion and Esser, Patrick and Kulal, Sumith and others},
  journal={arXiv preprint arXiv:2506.15742},
  year={2025}
}

@inproceedings{luo2024flowdiffuser,
  title={{FlowDiffuser}: Advancing optical flow estimation with diffusion models},
  author={Luo, Ao and Li, Xin and Yang, Fan and Liu, Jiangyu and Fan, Haoqiang and Liu, Shuaicheng},
  booktitle={CVPR},
  year={2024}
}

@inproceedings{saharia2022palette,
  title={Palette: Image-to-image diffusion models},
  author={Saharia, Chitwan and Chan, William and Chang, Huiwen and Lee, Chris and Ho, Jonathan and Salimans, Tim and Fleet, David and Norouzi, Mohammad},
  booktitle={SIGGRAPH},
  year={2022}
}

@article{saharia2022image,
  title={Image super-resolution via iterative refinement},
  author={Saharia, Chitwan and Ho, Jonathan and Chan, William and Salimans, Tim and Fleet, David J and Norouzi, Mohammad},
  journal={IEEE Transactions on Pattern Analysis and Machine Intelligence},
  volume={45},
  number={4},
  pages={4713--4726},
  year={2022},
}

@article{jiang2023low,
  title={Low-light image enhancement with wavelet-based diffusion models},
  author={Jiang, Hai and Luo, Ao and Fan, Haoqiang and Han, Songchen and Liu, Shuaicheng},
  journal={ACM Transactions on Graphics (TOG)},
  volume={42},
  number={6},
  pages={1--14},
  year={2023},
}

@inproceedings{wang2023exposurediffusion,
  title={{ExposureDiffusion}: Learning to expose for low-light image enhancement},
  author={Wang, Yufei and Yu, Yi and Yang, Wenhan and Guo, Lanqing and Chau, Lap-Pui and Kot, Alex C and Wen, Bihan},
  booktitle={ICCV},
  year={2023}
}

@article{kawar2022denoising,
  title={Denoising diffusion restoration models},
  author={Kawar, Bahjat and Elad, Michael and Ermon, Stefano and Song, Jiaming},
  journal={NeurIPS},
  volume={35},
  pages={23593--23606},
  year={2022}
}

@article{wang2022zero,
  title={Zero-shot image restoration using denoising diffusion null-space model},
  author={Wang, Yinhuai and Yu, Jiwen and Zhang, Jian},
  journal={arXiv preprint arXiv:2212.00490},
  year={2022}
}

@inproceedings{yuan2025generative,
  title={Generative photography: Scene-consistent camera control for realistic text-to-image synthesis},
  author={Yuan, Yu and Wang, Xijun and Sheng, Yichen and Chennuri, Prateek and Zhang, Xingguang and Chan, Stanley},
  booktitle={CVPR},
  year={2025}
}

@inproceedings{ren2025ispdiffuser,
  title={{ISPDiffuser}: Learning {RAW-to-sRGB} Mappings with Texture-Aware Diffusion Models and Histogram-Guided Color Consistency},
  author={Ren, Yang and Jiang, Hai and Yang, Menglong and Li, Wei and Liu, Shuaicheng},
  booktitle={AAAI},
  year={2025}
}

@article{chen2025rddm,
  title={{RDDM}: Practicing RAW Domain Diffusion Model for Real-world Image Restoration},
  author={Chen, Yan and Wen, Yi and Li, Wei and Liu, Junchao and Guo, Yong and Hu, Jie and Chen, Xinghao},
  journal={arXiv preprint arXiv:2508.19154},
  year={2025}
}

@inproceedings{reinders2025raw,
  title={Raw-diffusion: {RGB}-guided diffusion models for high-fidelity raw image generation},
  author={Reinders, Christoph and Berdan, Radu and Besbinar, Beril and Otsuka, Junji and Iso, Daisuke},
  booktitle={WACV},
  year={2025},
}

@inproceedings{wang2025lediff,
  title={{LEDiff}: Latent Exposure Diffusion for {HDR} Generation},
  author={Wang, Chao and Xia, Zhihao and Leimkuhler, Thomas and Myszkowski, Karol and Zhang, Xuaner},
  booktitle={CVPR},
  year={2025}
}

@article{bemana2025bracket,
  title={Bracket Diffusion: {HDR} Image Generation by Consistent {LDR} Denoising},
  author={Bemana, Mojtaba and Leimk{\"u}hler, Thomas and Myszkowski, Karol and Seidel, Hans-Peter and Ritschel, Tobias},
  journal={Computer Graphics Forum},
  pages={e70086:1--13},
  volume = {44},
  number = {2},
  year={2025},
}

@article{dai2023instructblip,
  title={{InstructBLIP}: Towards general-purpose vision-language models with instruction tuning},
  author={Dai, Wenliang and Li, Junnan and Li, Dongxu and Tiong, Anthony and Zhao, Junqi and Wang, Weisheng and Li, Boyang and Fung, Pascale N and Hoi, Steven},
  journal={NeurIPS},
  year={2023}
}

@inproceedings{abdelhamed2019noise,
  title={Noise flow: Noise modeling with conditional normalizing flows},
  author={Abdelhamed, Abdelrahman and Brubaker, Marcus A and Brown, Michael S},
  booktitle={ICCV},
  year={2019}
}

@article{foi2008practical,
  title={Practical {Poissonian-Gaussian} noise modeling and fitting for single-image raw-data},
  author={Foi, Alessandro and Trimeche, Mejdi and Katkovnik, Vladimir and Egiazarian, Karen},
  journal={IEEE Transactions on Image Processing},
  volume={17},
  number={10},
  pages={1737--1754},
  year={2008},
}

@inproceedings{abdelhamed2018high,
  title={A high-quality denoising dataset for smartphone cameras},
  author={Abdelhamed, Abdelrahman and Lin, Stephen and Brown, Michael S},
  booktitle={CVPR},
  year={2018}
}

@inproceedings{Gong2019ConvolutionalMA,
  title={Convolutional Mean: A Simple Convolutional Neural Network for Illuminant Estimation},
  author={Han Gong},
  booktitle={BMVC},
  year={2019},
}

@InProceedings{Punnappurath_2022_CVPR,
    author    = {Punnappurath, Abhijith and Abuolaim, Abdullah and Abdelhamed, Abdelrahman and Levinshtein, Alex and Brown, Michael S.},
    title     = {Day-to-Night Image Synthesis for Training Nighttime Neural {ISP}s},
    booktitle = {CVPR},
    year      = {2022},
}

@inproceedings{ronneberger2015u,
  title={{U-Net}: Convolutional Networks for Biomedical Image Segmentation},
  author={Ronneberger, Olaf and Fischer, Philipp and Brox, Thomas},
  booktitle={MICCAI},  
  year={2015},
}

@inproceedings{zamir2022restormer,
  title={Restormer: Efficient transformer for high-resolution image restoration},
  author={Zamir, Syed Waqas and Arora, Aditya and Khan, Salman and Hayat, Munawar and Khan, Fahad Shahbaz and Yang, Ming-Hsuan},
  booktitle={CVPR},
  year={2022}
}

@inproceedings{ignatov2020replacing,
  title={Replacing mobile camera {ISP} with a single deep learning model},
  author={Ignatov, Andrey and Van Gool, Luc and Timofte, Radu},
  booktitle={CVPRW},
  year={2020}
}

@InProceedings{Kim_2024_CVPR,
    author    = {Kim, Woohyeok and Kim, Geonu and Lee, Junyong and Lee, Seungyong and Baek, Seung-Hwan and Cho, Sunghyun},
    title     = {{ParamISP}: Learned Forward and Inverse {ISPs} using Camera Parameters},
    booktitle = {CVPR},
    year      = {2024},
}

@inproceedings{afifi2025time,
  title={Time-aware auto white balance in mobile photography},
  author={Afifi, Mahmoud and Zhao, Luxi and Punnappurath, Abhijith and Abdelsalam, Mohamed A and Zhang, Ran and Brown, Michael S},
  booktitle={ICCV},
  year={2025}
}

@article{afifi2025modular,
  title={Modular Neural Image Signal Processing},
  author={Afifi, Mahmoud and Wang, Zhongling and Zhang, Ran and Brown, Michael S},
  journal={arXiv preprint arXiv:2512.08564},
  year={2025}
}

@inproceedings{zhang2021learning,
  title={Learning raw-to-s{RGB} mappings with inaccurately aligned supervision},
  author={Zhang, Zhilu and Wang, Haolin and Liu, Ming and Wang, Ruohao and Zhang, Jiawei and Zuo, Wangmeng},
  booktitle={ICCV},
  year={2021}
}

@inproceedings{afifi2021semi,
  title={Semi-supervised raw-to-raw mapping},
  author={Afifi, Mahmoud and Abuolaim, Abdullah},
  booktitle={BMVC},
  year={2021}
}

@inproceedings{perevozchikov2024rawformer,
  title={Rawformer: Unpaired raw-to-raw translation for learnable camera {ISPs}},
  author={Perevozchikov, Georgy and Mehta, Nancy and Afifi, Mahmoud and Timofte, Radu},
  booktitle={ECCV},
  year={2024},
}

@article{delbracio2021mobile,
  title={Mobile computational photography: A tour},
  author={Delbracio, Mauricio and Kelly, Damien and Brown, Michael S and Milanfar, Peyman},
  journal={Annual review of vision science},
  volume={7},
  number={1},
  pages={571--604},
  year={2021},
}

@inproceedings{debevec2008recovering,
  title={Recovering high dynamic range radiance maps from photographs},
  author={Paul E. Debevec and Jitendra Malik},
  booktitle={ACM SIGGRAPH},
  year={2008}
}

@INPROCEEDINGS{mitsunaga1999radiometric, 
 author={Mitsunaga, Tomoo and Nayar, Shree K.},
 booktitle={CVPR}, 
title={Radiometric self calibration}, 
year={1999}, 
}

@article{grossberg2003determining,
 author = {Grossberg, Michael D. and Nayar, Shree K.},
 title = {Determining the Camera Response from Images: What Is Knowable?},
 journal = {IEEE Transactions on Pattern Analysis and Machine Intelligence},
 volume = {25},
 number = {11},
 year = {2003},
 issn = {0162-8828},
 pages = {1455--1467},
 numpages = {13},
}

@article{chakrabarti2014modeling,
  title={Modeling radiometric uncertainty for vision with tone-mapped color images},
  author={Chakrabarti, Ayan and Xiong, Ying and Sun, Baochen and Darrell, Trevor and Scharstein, Daniel and Zickler, Todd and Saenko, Kate},
  journal={IEEE Transactions on Pattern Analysis and Machine Intelligence},
  volume={36},
  number={11},
  pages={2185--2198},
  year={2014},
}

@inproceedings{Chakrabarti2009empirical,
  title={An Empirical Camera Model for Internet Color Vision},
  author={Ayan Chakrabarti and Daniel Scharstein and Todd E. Zickler},
  booktitle={BMVC},
  year={2009}
}

@inproceedings{nam2017modelling,
  title={Modelling the scene dependent imaging in cameras with a deep neural network},
  author={Nam, Seonghyeon and Joo Kim, Seon},
  booktitle={ICCV},
  year={2017}
}

@inproceedings{conde2022model,
  title={Model-based image signal processors via learnable dictionaries},
  author={Conde, Marcos V and McDonagh, Steven and Maggioni, Matteo and Leonardis, Ales and P{\'e}rez-Pellitero, Eduardo},
  booktitle={AAAI},
  year={2022}
}

@article{kim2012new,
  title={A new in-camera imaging model for color computer vision and its application},
  author={Kim, Seon Joo and Lin, Hai Ting and Lu, Zheng and S{\"u}sstrunk, Sabine and Lin, Stephen and Brown, Michael S.},
  journal={IEEE Transactions on Pattern Analysis and Machine Intelligence},
  volume={34},
  number={12},
  pages={2289--2302},
  year={2012},
}

@inproceedings{lan,
  title={{LAN}: Lightweight attention-based network for raw-to-{RGB} smartphone image processing},
  author={Raimundo, Daniel Wirzberger and Ignatov, Andrey and Timofte, Radu},
  booktitle={CVPRW},
  year={2022}
}

@inproceedings{microisp,
  title={{MicroISP}: Processing {32MP} Photos on Mobile Devices with Deep Learning},
  author={Ignatov, Andrey and Sycheva, Anastasia and Timofte, Radu and Tseng, Yu and Xu, Yu-Syuan and Yu, Po-Hsiang and Chiang, Cheng-Ming and Kuo, Hsien-Kai and Chen, Min-Hung and Cheng, Chia-Ming and others},
  booktitle={ECCV},
  year={2022},
}

@inproceedings{fourier,
  title={Enhancing RAW-to-s{RGB} with decoupled style structure in {F}ourier domain},
  author={He, Xuanhua and Hu, Tao and Wang, Guoli and Wang, Zejin and Wang, Run and Zhang, Qian and Yan, Keyu and Chen, Ziyi and Li, Rui and Xie, Chengjun and others},
  booktitle={AAAI},
  year={2024}
}

@InProceedings{dit,
    author    = {Peebles, William and Xie, Saining},
    title     = {Scalable Diffusion Models with Transformers},
    booktitle = {ICCV},
    year      = {2023},
}

@inproceedings{
hu2022lora,
title={Lo{RA}: Low-Rank Adaptation of Large Language Models},
author={Edward J Hu and Yelong Shen and Phillip Wallis and Zeyuan Allen-Zhu and Yuanzhi Li and Shean Wang and Lu Wang and Weizhu Chen},
booktitle={ICLR},
year={2022},
}

@Article{Maaten_2008_JMLR,
    author    = {van der Maaten, Laurens and Hinton, Geoffrey},
    title     = {Visualizing Data using t-SNE},
    journal   = {Journal of Machine Learning Research},
    volume    = {9},
    pages     = {2579--2605},
    year      = {2008}
}

@Article{McInnes_2018_UMAP,
    author    = {McInnes, Leland and Healy, John and Melville, James},
    title     = {UMAP: Uniform Manifold Approximation and Projection for Dimension Reduction},
    journal   = {arXiv preprint arXiv:1802.03426},
    year      = {2018}
}

@Article{Pearson_1901_PCA,
    author    = {Pearson, Karl},
    title     = {On Lines and Planes of Closest Fit to Systems of Points in Space},
    journal   = {Philosophical Magazine},
    volume    = {2},
    number    = {11},
    pages     = {559--572},
    year      = {1901}
}

@inproceedings{fivek,
	author = "Vladimir Bychkovsky and Sylvain Paris and Eric Chan and Fr{\'e}do Durand",
	title = "Learning Photographic Global Tonal Adjustment with a Database of Input / Output Image Pairs",
	booktitle = CVPR,
	year = {2011}
}
\end{document}